%% file: main.tex
\documentclass[10pt,twocolumn,letterpaper]{article}

\usepackage{cvpr}              % To produce the CAMERA-READY version
\input{preamble}
\definecolor{cvprblue}{rgb}{0.21,0.49,0.74}
\def\paperID{*******} % *** Enter the Paper ID here
\def\confName{CVPR}
\def\confYear{2026}

\title{S$^{2}$FT: Parameter-Efficient Fine-Tuning in Sparse Spectrum Domain}

\author{\normalsize Baoquan Zhang$^1$, Zhehao Yu$^1$, Lisai Zhang$^3$, Kenghong Lin$^{*1}$, Tianran Chen$^{1}$, Yuxi Sun$^{4}$, Yunming Ye\thanks{Corresponding author.}$\ ^{1}$, Yao He$^2$\\
	\normalsize $^1$ Harbin Institute of Technology, Shenzhen; $^2$ ShenZhen SiFar Co., Ltd.; $^3$ Bilibili. Inc; $^4$ Shenzhen University \\
	{\tt\small baoquanzhang@hit.edu.cn, \{210110629, linkenghong\}@stu.hit.edu.cn, lisaizhang@foxmail.com, }\\
	{\tt\small 23S151121@stu.hit.edu.cn, sunyuxi@szu.edu.cn, yeyunming@hit.edu.cn, heyao18818@gmail.com}\\
    }

\begin{document}
\maketitle
\input{sec/0_abstract}    
\input{sec/1_intro}
\input{sec/2_relatedwork}
\input{sec/3_method}
\input{sec/4_experiment}
\input{sec/5_conclusion}

{
    \small
    \bibliographystyle{ieeenat_fullname}
    \bibliography{main}
}

%     \small
%     \bibliographystyle{ieeenat_fullname}
%     \bibliography{main}
% }
\end{document}

%% file: sec/0_abstract.tex
\begin{abstract}
Parameter Efficient Fine-Tuning (PEFT) is a key technique for adapting a large pretrained model to downstream tasks by fine-tuning only a small number of parameters. 
Recent methods based on Fourier transforms have further reduced the fine-tuned parameters scale by only fine-tuning a few spectral coefficients. Its basic assumption is that the weight change $\Delta W$ is a spatial-domain matrix with a sparse spectrum. 
However, in this paper, we observe that the spectrum of weight change is not sparse, but instead distributed like power-uniform. This fact implies that fine-tuning only a few spectral coefficients is insufficient to accurately model the weight change with uniform spectrum. 
To address this issue, we propose to seek an invertible transformation that can transform a latent spatial-domain matrix with sparse spectrum to the weight change, and then perform PEFT on such sparse spectrum domain with few spectral coefficients, called S$^2$FT.
To seek such transformation, we first pre-estimate a coarse weight change as a prior.  
Then, inspired by that sparse spectrum often correspond to locally smooth spatial structures, we regard this transformation as a row and column rearrangement operation on the pre-estimated weight change that smooth spatial structures while keep the structure information of neurons. 
Finally, we propose to solve the rearrangement search problem in a simple nearest neighbor search manner, thereby obtaining the invertible transformation.
Extensive results show our S$^2$FT achieves superior performance by only using 0.08\% training parameters.
\end{abstract}

%% file: sec/1_intro.tex
\section{Introduction}

\label{sec:intro}
Pretrained large models have gained significant attention in computer vision and natural language processing \cite{he2022masked, kirillov2023segment, dosovitskiy2020image, zhang2025zookt, zhang2024codebook, lin2023vision}. In earlier works, full fine-tuning is commonly used to adapt these pretrained models to specific downstream tasks. However, as the model size increases, the computational resources required for full fine-tuning and parameter storage become a limiting factor. To address this challenge, Parameter Efficient Fine-Tuning (PEFT) was introduced recently and has received widespread interest, which aims to adapt a pretrained large model to downstream tasks by fine-tuning only few trainable parameters \cite{hulora, he2023sensitivity, zhang2024gradient, liudora, zhu2024asymmetry}.
 
\begin{figure}[t]
  \centering

  \begin{subfigure}{0.99\linewidth}
    \centering
    \includegraphics[width=\linewidth]{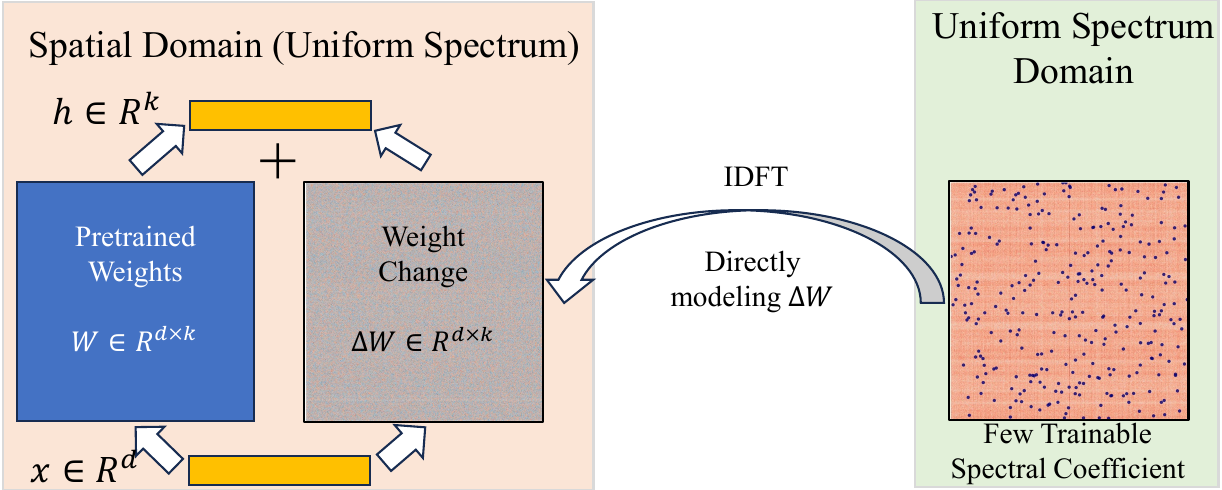}
    \caption{Existing FourierFT.}\label{fig1_a}
  \end{subfigure}

  \par\vspace{0.6em}

  \begin{subfigure}{0.99\linewidth}
    \centering
    \includegraphics[width=\linewidth]{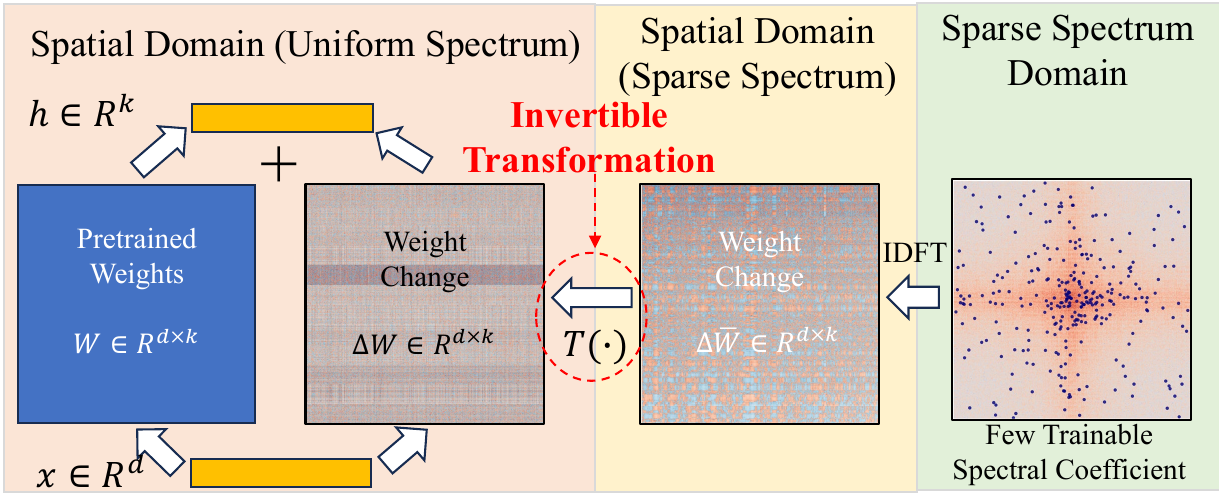}
    \caption{Our S$^2$FT.}\label{fig1_b}
  \end{subfigure}

  \caption{Existing FourierFT (a) directly models weight change $\Delta W$ by few trainable spectral coefficients, the performance is limited by the non-frequency-sparse nature of $\Delta W$. Our S$^2$FT (b) seeks an invertible transformation from a latent spatial-domain matrix $\Delta \bar{W}$ with sparse spectrum to the $\Delta W$, and perform PEFT on the sparse spectrum domain with fewer spectral coefficients. (In the rightmost plot, the orange depth indicates the power spectrum, and blue dots denote trainable spectral coefficients.)}
  \vspace{-15pt}
  \label{fig1}
\end{figure}

Low-Rank Adaptation (LoRA) \cite{hulora} is a popular method for PEFT, which aims to model the weight change $\Delta W$ by using a product between two low-rank matrices $A$ and $B$ as $\Delta W=AB$. The parameter scale of the low-rank matrix is much smaller than that of the original parameters. Despite LoRA’s superior performance, the parameter scale for large models is still heavy, imposing storage burden for both public communities and individual users.
Recently, Gao et al. \cite{gao2024parameter} further decrease the parameter scale by a novel Fourier transform-based method called FourierFT. As shown in Figure~\ref{fig1_a}, it directly regards $\Delta W$ as a spatial-domain matrix, and then achieves PEFT by fine-tuning on spectral domain with a few trainable spectral coefficients. The basic assumption of such design is that the weight change $\Delta W$ is a spatial domain matrix with a sparse spectrum. However, a natural question is that \emph{is such assumption really reasonable?}

% To answer this question, in this paper, we  conduct a detailed analysis on the weight change $\Delta W$ and find that when directly regarding the weight change $\Delta W$ as spatial-domain matrix, its spectrum is usually not sparse but shows a format of the power-uniform spectrum (see Section~\ref{section32} for details). In such an uniform spectrum distribution, it is difficult to select only a few spectral coefficients to recover the original weigh change, which limits PEFT performance.
% Ideally, FourierFT works better for smooth $\Delta W$ with a sparse spectrum, which can be approximated precisely with few coefficients. 

To answer this question, we conduct a detailed analysis of the weight change $\Delta W$, and observe that when it is directly treated as a spatial-domain matrix, its spectrum is typically not sparse but instead exhibits a power-uniform distribution (see Section~\ref{section32} for details). Under such a distribution, it is difficult to identify a small number of spectral coefficients that can accurately reconstruct the original weight change, which limits PEFT's effectiveness. Ideally, Fourier-based PEFT perform best when the underlying $\Delta W$ is smooth in spatial domain, corresponding to a sparse spectrum that can be well-approximated using only a few spectral coefficients.

Motivated by this, we propose a novel PEFT approach in the sparse spectral domain, termed S$^2$FT. As shown in Figure~\ref{fig1_b}, instead of 
% directly treating the weight change $\Delta W$ with a uniform spectrum as the spatial-domain matrix, 
perform PEFT on the $\Delta W$'s  uniform spectrum, 
we propose to seek an invertible transformation that maps a smooth latent  matrix $\Delta \bar{W}$ —whose spectrum is sparse—into $\Delta W$ and
% then perform PEFT in its sparse domain by optimizing only a few spectral coefficients. 
then perform PEFT on the spectrum-sparse latent matrix $\Delta \bar{W}$ by optimizing only a few of its spectral coefficients.
The key advantage of this approach lies in its ability to fully leverage the expressive power of a sparse spectrum and accurately model $\Delta W$ with fewer trainable parameters. 
The key challenge is to discover an appropriate invertible transformation. To tackle this, we first pre-estimate a coarse weight change $\Delta \hat{W}$ via gradient accumulation on a small training subset, serving as a prior. 
Then, inspired by the well-known principle that sparse spectra often correspond to locally smooth spatial structures, we formulate the invertible transformation as a solution of row and column rearrangement problem on $\Delta \hat{W}$ that minimize the distance between adjacent rows and columns. The rearrangement should smooth the weight matrix while do not disrupt the structure information of the neurons. 
Finally, we propose to solve the rearrangement
problem in a simple nearest neighbor search manner, thereby obtaining the invertible transformation for achieving our S$^2$FT.

Our main contributions can be summarized as follows:
\begin{itemize}
   \item We experimentally confirm that the spectrum of weight change is not sparse but rather tends to be power-uniform. This indicates that fine-tuning only a few spectral coefficients is insufficient for accurately modeling $\Delta W$, which limits the performance of PEFT. To the best of our knowledge, this is the first work to identify this challenge.
   \item We propose a novel PEFT method in sparse spectrum domain, which seeks an invertible transformation that can transform a latent spatial-domain matrix with sparse spectrum to the weight change, and then perform PEFT on such sparse spectrum domain. 
   Its advantage is that a few spectral coefficients can be fully exploited for PEFT.
   %Its advantage is that the spectral coefficients is more effective for PEFT.
   \item We conduct experiments on various tasks and datasets, which verify the effectiveness of our S$^2$FT.
\end{itemize}

%% file: sec/2_relatedwork.tex
\section{Related Works}
\label{sec:formatting}
\subsection{Parameter-Efficient Fine-Tuning (PEFT)}  
PEFT \cite{mercea2024time} is a challenging task, which aims to tune only few trainable parameters to achieve adaptation of pretrained models on downstream tasks. Existing methods can be roughly categorized into three groups: 
\textbf{(1) Reparameterization-based Methods.} These methods primarily center on directly modeling the weight changes using few trainable parameters and executing PEFT through reparameterization techniques\cite{hulora, guo2024learning, kopiczkovera, zhaogalore, lian2022scaling, chen2024quanta, NEURIPS2024_75008a0f, huang2025hira, liao2025hmora}. For instance, %Hu et al. \cite{hulora} introduced a Low-Rank Adaptation approach, termed LoRA, which models weight changes as the product of two low-rank matrices. Similarly, 
Gao et al. \cite{gao2024parameter} employed a Fourier-domain transformation, referred to as FourierFT, under the assumption that weight changes can be represented as a spatial-domain matrix with a sparse spectrum. This enables PEFT by fine-tuning only a select few spectral coefficients.
\textbf{(2) Addition-based Methods.} This class of methods achieves PEFT by integrating additional adapters or optimizing prompts and prefixes within pre-trained layers \cite{chen2024conv, peng2024parameter, han20232, yin20231, tu2023visual, houlsby2019parameter}. %Gao et al. \cite{houlsby2019parameter} exemplified this by embedding lightweight neural modules into the pre-trained base model layers. These modules are designed to learn a task-specific feature adapter, thereby facilitating PEFT.
\textbf{(3) Selection-based Methods}. This category emphasizes tuning only a subset of the pre-trained parameters to achieve PEFT \cite{he2023sensitivity, zaken2022bitfit, houlsby2019parameter, zhang2024gradient, he2025smt, woo2025paca}. 
% For example, He et al. \cite{he2023sensitivity} presented a sensitivity-aware PEFT method, named SPT, which identifies a subset of parameters based on their sensitivity. %Subsequently, PEFT is performed through a combination of unstructured and structured tuning strategies.
%In this paper, we focus on Fourier transformation-based PEFT methods. Differently, we discover a key challenge  overlooked in previous studies, i.e., the power spectrum of weight change $\Delta W$ is not sparse, but tends to uniform. Accordingly, we present a new PEFT method in sparse spectrum domain, which aims to resorting to its spectrum-sparse character to enhance weight change modeling with few spectral coefficients.
In this paper, we focus on Fourier transformation-based methods. Differently, instead of directly fine-tuning coefficients on a uniform spectrum domain, we propose to perform fine-tuning on a sparse spectrum domain so that fewer trainable coefficients are required for further enhancing PEFT performance.  
% we identify a key challenge  overlooked in previous studies, i.e., the power spectrum of weight change  $\Delta W$ is not sparse but tends to be uniformly distributed. To address this issue, we propose a novel PEFT method in the sparse spectral domain, leveraging its inherent spectral sparsity to enhance weight change modeling using only a few spectral coefficients.

\begin{figure*}[t]
  \centering

  \begin{subfigure}[t]{0.24\linewidth}
    \centering
    \includegraphics[width=\linewidth]{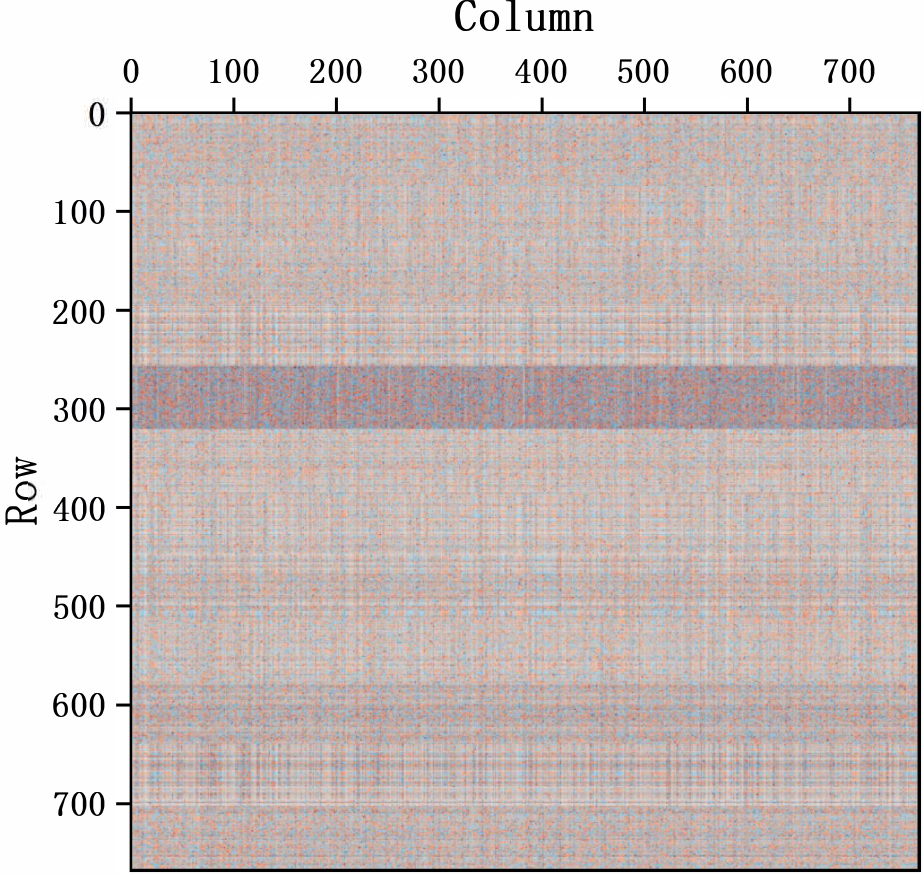}
    \caption{Weight change $\Delta W$.}\label{fig2_a}
  \end{subfigure}\hfill
  \begin{subfigure}[t]{0.24\linewidth}
    \centering
    \includegraphics[width=\linewidth]{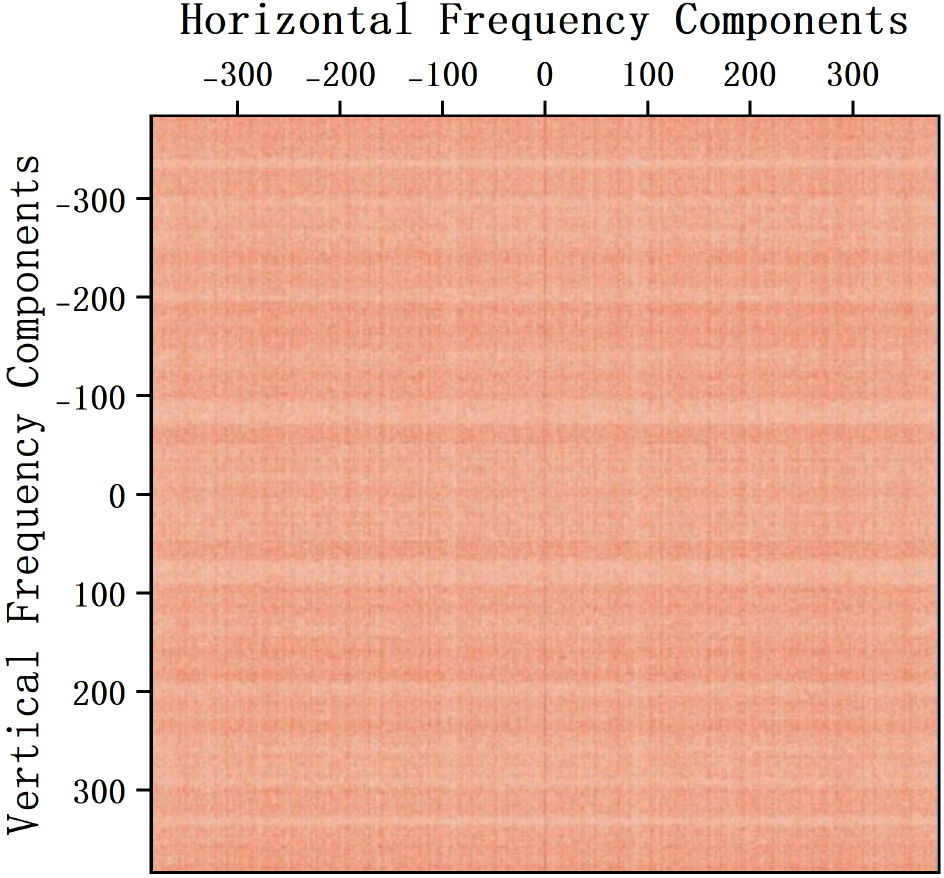}
    \caption{Power spectrum of $\Delta W$.}\label{fig2_b}
  \end{subfigure}\hfill
  \begin{subfigure}[t]{0.24\linewidth}
    \centering
    \includegraphics[width=\linewidth]{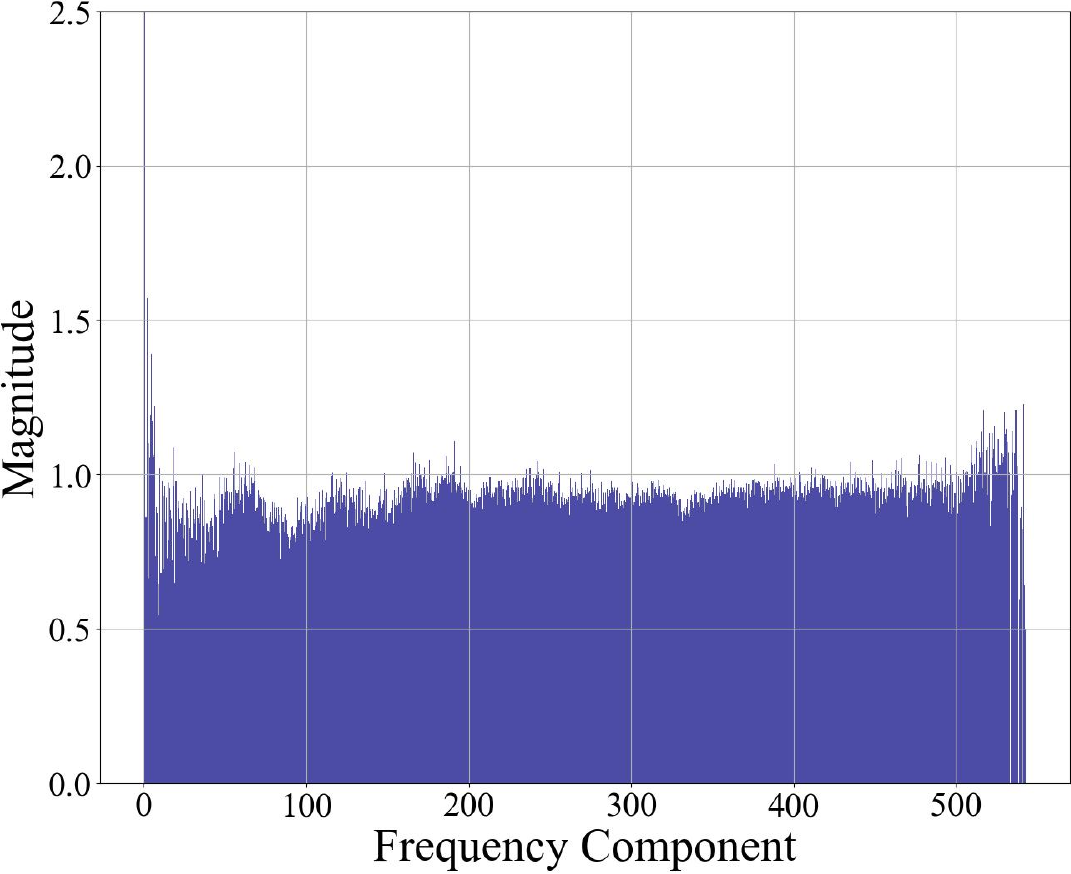}
    \caption{Amplitude distribution.}\label{fig2_c}
  \end{subfigure}\hfill
  \begin{subfigure}[t]{0.24\linewidth}
    \centering
    \includegraphics[width=\linewidth]{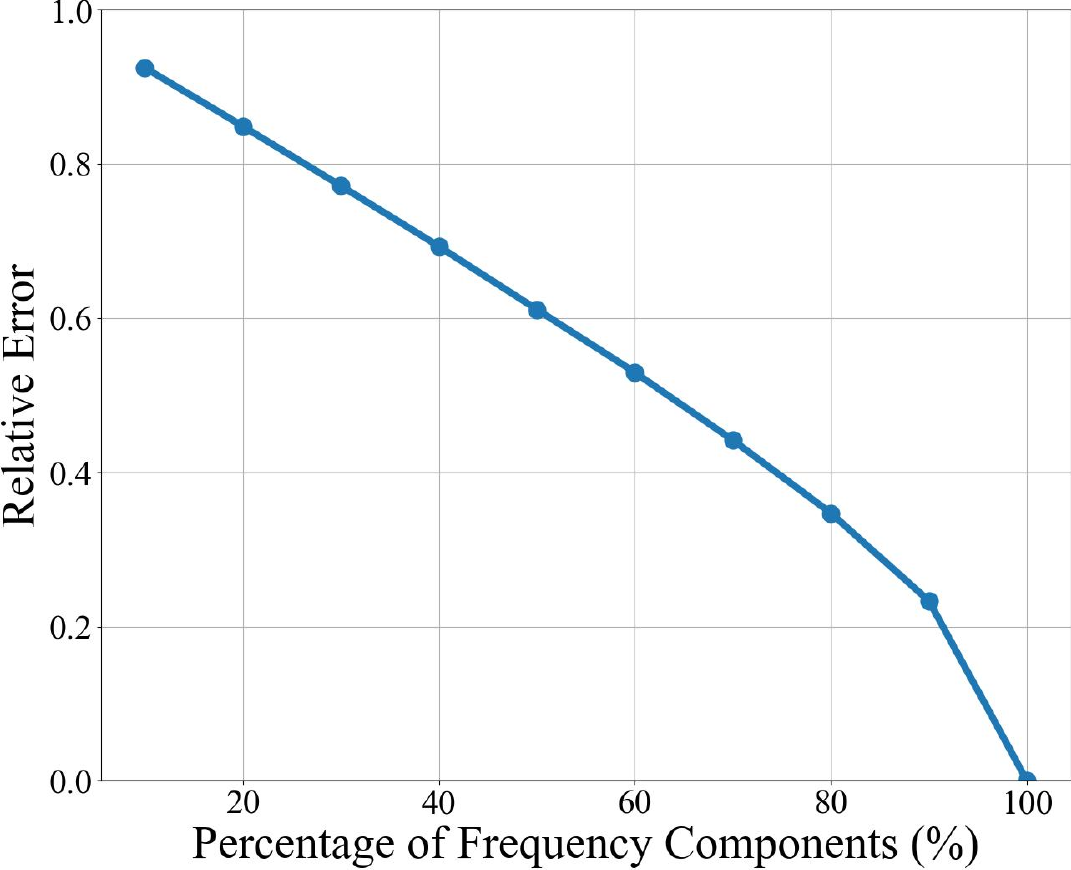}
    \caption{Relative error of $\Delta W$.}\label{fig2_d}
  \end{subfigure}
  \vspace{-5pt}
  \caption{Distribution analysis of weight change $\Delta W$ in spatial and spectral domains.}
  \vspace{-10pt}
  \label{fig2}
\end{figure*}

\subsection{Fourier Transform for Deep Learning} 
Fourier transform plays a fundamental role across information processing, providing an effective way to extract frequency components. Fourier transform has been used for representation learning~\cite{fft_1,fft_2,fft_3,fft_4}.
For example, in \cite{conrad2017sparse, mevenkamp2016variational}, Fourier transform is used to convert spatial-domain information into frequency representations to extract discriminative patterns from challenging datasets containing noise or high dimensionality.
Recently, Gao et al.~\cite{gao2024parameter} propose viewing the weight change matrix $\Delta W$ as a spatial-domain matrix and performing PEFT by applying a Fourier transform, thereby fine-tuning only a small number of its spectral coefficients. However, in this paper, we observe that the spectrum of the weight change is not inherently sparse, but instead exhibits a power-uniform distribution. To address this, we introduce a latent spatial-domain matrix that yields a sparse spectrum through an invertible transformation. By fine-tuning the spectral coefficients derived from this sparsified representation, our method enables more effective PEFT.

%% file: sec/3_method.tex
\section{Methodology}
\subsection{Preliminaries and Motivation Analysis}
\subsubsection{Preliminaries: FourierFT}
Formally, let $W$ denotes a pre-trained weight matrix, i.e., $W \in \mathbb{R}^{d \times k}$ and $\Delta W$ is its weight change after adapting to downstream tasks. The key challenge of achieving PEFT is how to model the $\Delta W$ by using only few trainable parameters. Recently, Gao et al. \cite{gao2024parameter} effectively address this challenge by a Fourier transform-based method (FourierFT). Specifically, they firstly regard the weight change $\Delta W$ as a spatial-domain matrix. Then, they randomly initialize a spectral-domain matrix $F \in \mathbb{C}^{d \times k}$ where $\mathbb{C}$ denotes complex field and only few spectral coefficients are nonzero (i.e., trainable parameters) and other are zero. After that, the weight change $\Delta W$ can be represented as a spatial-domain matrix obtained by using an inverse Fourier transform on spectral-domain matrix $F \in \mathbb{C}^{d \times k}$:
\begin{equation}
    \Delta W = \mathcal{R}(IDFT(F))*\alpha,
\end{equation}
where $\mathcal{R}(\cdot)$ denotes an operation taking the real part of the complex matrix, $IDFT(\cdot)$ is the inverse Fourier transform, and $\alpha$ is a hyper-parameter. Finally, the adapted weight $W'$ can be represented as:
\begin{equation}
    W' = W + \Delta W = W + \mathcal{R}(IDFT(F))*\alpha. 
\end{equation}
The assumption of leveraging few spectral coefficients to model the weight change $\Delta W$ is that its spectrum is very sparse. However, a natural question is that \emph{is such assumption really reasonable?}

% \begin{figure}
%   \centering

%   \subfigure[Weight change $\Delta W$.]{
%     \begin{minipage}[t]{0.45\linewidth}
%       \centering
%       \includegraphics[width=1.0\linewidth]{sec/fig2_a.pdf}
%     \end{minipage}
%     \label{fig2_a}
%   }
%   \subfigure[Power spectrum $\Delta W$.]{
%     \begin{minipage}[t]{0.45\linewidth}
%       \centering
%       \includegraphics[width=1.0\linewidth]{sec/fig2_b.pdf}
%     \end{minipage}
%     \label{fig2_b}
%   }
%   \\
%   \subfigure[Amplitude distribution.]{
%     \begin{minipage}[t]{0.45\linewidth}
%       \centering
%       \includegraphics[width=1.0\linewidth]{sec/magdis_01.pdf}
%     \end{minipage}
%     \label{fig2_c}
%   }
%   \subfigure[Relative error of $\Delta W$.]{
%     \begin{minipage}[t]{0.45\linewidth}
%       \centering
%       \includegraphics[width=1.0\linewidth]{sec/err_1.pdf}
%     \end{minipage}
%     \label{fig2_d}
%   }

%   \caption{Distribution analysis of weight change $\Delta W$ in spatial domain and spectral domain.}
%   \label{fig2}
% \end{figure}

\subsubsection{Motivation Analysis on Weight Change $\triangle W$}
\label{section32}
To answer the above question,  we conduct a detailed analysis on weight change $\triangle W$. Specifically, we randomly select a weight matrix $W$ from a pretrained vision transformer (ViT) model and leverage full fine-tuning to calculate its weight change $\triangle W$. Then, we visualize $\triangle W$ and its power spectrum in Figures~\ref{fig2_a} and \ref{fig2_b} (please see Appendix for more results). It can be clearly observed that (1) the distribution of $\triangle W$ is chaotic in spatial domain (see Figure~\ref{fig2_a}), which means that it contains various frequency components; (2) the power distribution is close to uniform in the entire spectral domain (see Figure~\ref{fig2_b}). Furthermore, we visualize the amplitude at different frequency components in Figure~\ref{fig2_c}. It can be  found that the power spectrum of weight change $\Delta W$ is indeed not sparse, but tends to be power-uniform. This suggests that modeling  $\triangle W$  using few spectral coefficients is very difficult, which limits the PEFT performance of FourierFT. To further verify this observation, we visualize the average relative reconstruction error of every $\triangle W$ in ViT  in Figure~\ref{fig2_d} where we gradually sample its spectral coefficients from 10\% to 100\%. We notice that it requires to sample over 90\% spectral coefficients to achieve less than 10\% reconstruction error. 

The above analysis indicates a fact that directly using few spectral coefficients is insufficient to model the weight change $\Delta W$ since it is a spectrum-uniform matrix. Thus, an intuitive insight is \emph{whether there is a potential spatial-domain matrix $\Delta \bar{W}$ with sparse spectrum which can be non-destructively transformed to the weight change $\Delta W$?} If it exists, we can perform PEFT on this sparse spectrum such that few spectral coefficients can be fully exploited for modeling the weight change $\Delta W$.

\subsection{PEFT in Sparse Spectrum Domain (S$^2$FT)}
\label{sec_34}
Based on the above insight, we propose a novel parameter-efficient fine-tuning method in sparse spectrum domain  as shown in Figure~\ref{fig1_b}, called S$^2$FT. Instead of directly modeling the weight change $\Delta W$ with an uniform spectrum, we attempt to 
seek an invertible transformation that can transform a latent spatial-domain matrix $\Delta \bar{W}$ with sparse spectrum to the weight change $\Delta W$, and then perform PEFT in such sparse spectrum space with few spectral coefficients. %The advantage of such design is that few spectral coefficients can be fully exploited for accurately modeling weight change $\Delta W$. 
To formulate, given a pretrained weight $W$ and a downstream task $\tau=\{\mathcal{D}_{tr}, \mathcal{D}_{te}\}$ where $\mathcal{D}_{tr}$ and $\mathcal{D}_{te}$ denote its training and test set, we first pre-estimate a coarse weight change $\Delta \hat{W}$ for each weight $W$ as a prior. Then, we leverage the prior $\Delta \hat{W}$ to seek an invertible transformation that can transform a latent spatial-domain matrix $\Delta \bar{W}$ with sparse spectrum to the weight change $\Delta W$. Finally, we perform PEFT in such sparse spectrum domain, where the adapted weight $W'$ on downstream tasks can be expressed as:
\begin{equation}
\begin{aligned}
    % W' = W + \Delta W = W + T(\Delta \bar{W}) = W + T(\mathcal{R}(IDFT(F))*\alpha). 
     W' &= W + \Delta W \\&= W + T(\Delta \bar{W})\\&=W + T(\mathcal{R}(IDFT(F))*\alpha). 
    \end{aligned}
    \label{eq_3}
\end{equation}
where $T(\cdot)$ denotes the invertible transformation from a latent spatial-domain matrix $\Delta \bar{W}$ with sparse spectrum to the weight change $\Delta W$; $F$ is a sparse complex matrices, i.e., only few values are learnable coefficients and others are zeros (see Section~\ref{sec_4_3} for its allocation details). Next, we introduce the details of the weight change pre-estimation, invertible transformation, and spectral coefficient sampling strategy on Sections~\ref{sec_4_1}, \ref{sec_4_2}, and \ref{sec_4_3}, respectively.

\subsubsection{Weight Change Pre-estimation}
\label{sec_4_1}
In order to obtain the proper transformation $T(\cdot)$, we need to pre-estimate the pattern of $\Delta W$ first and then leverage it to guide us to seek the invertible transformation $T(\cdot)$. 
An ideal estimation way is first leveraging full fine-tuning to obtain the adapted weight $W'$ with the entire training set $\mathcal{D}_{tr}$ and calculate the weight change $\Delta \hat{W}$ by using their difference, i.e., $\Delta \hat{W} = W'-W$. However, such method is 1) infeasible since it consumes huge computing cost; and 2) is contradictory to PEFT since the goal of PEFT is to reduce the full fine-tuning's training cost. 
Interestingly, we find that an very accurate estimation of $\Delta W$ is not required; actually, even a coarse approximation can also effectively guide the search for this invertible transformation $T(\cdot)$ (see Table 2 in Appendix). 
To this end, we propose a simple yet efficient pre-estimation method that leveraging the negative cumulative gradient of a subset $\mathcal{D}_{sub}$ from training set $\mathcal{D}_{tr}$ to pre-estimate a coarse weight change $\Delta \hat{W}$:
\begin{equation}
    \Delta \hat{W} = - \sum_{x \in \mathcal{D}_{sub}}{\nabla \mathcal{L}_{x}(w_i)},
    \label{eq_4}
\end{equation}
where $\nabla \mathcal{L}_{x}(W)$ is the gradient of weight $W$ from training sample $x$. 
We note that some PEFT works also calculate gradient over all training data \cite{he2023sensitivity, zhang2024gradient}, however, different from these methods that leverage gradient to estimate weight importance, we leverage it to pre-estimate weight change.

%Note that while some prior works also compute gradients over the training data to estimate weight importance \cite{choromanska2015loss, sanh2020movement}, we instead leverage them to approximate the weight change, further supporting the feasibility of our approach.

% \noindent{\textbf{Theoretical Analysis}:} \emph{why can negative cumulative gradient pre-estimate the weight change $\Delta \hat{W}$?}
% Such estimation is reasonable in theory because we usually use gradient descent algorithm to update the model parameters. The process can be expressed as:
% \begin{equation}
%     W' = W - \eta \sum_{x \in \mathcal{D}_{sub}}{\nabla \mathcal{L}_{x}(w_i)}.
% \end{equation}
% That is, $\Delta \hat{W} = W'-W = - \eta \sum_{x \in \mathcal{D}_{sub}}{\nabla \mathcal{L}_{x}(w_i)}$. Note that $\eta$ is learning rate, which is constant and can be overlooked. Therefore, Eq.~\eqref{eq_4} can be regarded 

\textbf{\emph{Why can the negative cumulative gradient roughly pre-estimate the weight change $\Delta \hat{W}$?}} 
In fact, the weight change $\Delta \hat{W}$ can be viewed as the direction of model weight update, i.e. moving the pretrained weights along such direction can minimize the loss on the training samples of a downstream task. 
According to the theory of gradient descent, we know that updating model parameters along the opposite direction of the cumulative parameter gradient can effectively reduce the model loss on training samples. 
This means that the opposite direction of cumulative gradient on training samples is exactly the weight update direction we are looking for. Thus, the negative cumulative gradient can be used to roughly pre-estimate the weight change $\Delta \hat{W}$. %This means that the negative cumulative gradient on training samples is exactly a pre-estimatation of weight change $\Delta \hat{W}$.
% Such estimation is reasonable in theory because  we usually use gradient descent algorithm to update the model parameters. The process can be expressed as:
% \begin{equation}
%     W' = W - \eta \sum_{x \in \mathcal{D}_{sub}}{\nabla \mathcal{L}_{x}(w_i)}.
% \end{equation}
% That is, $\Delta \hat{W} = W'-W = - \eta \sum_{x \in \mathcal{D}_{sub}}{\nabla \mathcal{L}_{x}(w_i)}$. Note that $\eta$ is learning rate, which is constant and can be overlooked. Therefore, Eq.~\eqref{eq_4} can be regarded 

% Inspired by the stochastic gradient descent process of full fine-tuning, we propose to randomly selecting a subset $\mathcal{D}_{sub}$ from training set $\mathcal{D}_{tr}$ and leveraging its cumulative gradient to pre-estimate the weight change $\Delta \hat{W}$. That is,
% \begin{equation}
%     \Delta \hat{W} = - \sum_{x \in \mathcal{D}_{sub}}{\nabla \mathcal{L}_{x}(w_i)},
%     \label{eq_4}
% \end{equation}
% where $\nabla \mathcal{L}_{x}(W)$ denotes the gradient of weight $W$ from training sample $x$. Thus, the process of performing one-step full fine-tuning can be expressed as:
% \begin{equation}
%     W' = W - \eta \sum_{x \in \mathcal{D}_{sub}}{\nabla \mathcal{L}_{x}(w_i)}.
% \end{equation}
% Therefore, $\Delta \hat{W} = W'-W = - \eta \sum_{x \in \mathcal{D}_{sub}}{\nabla \mathcal{L}_{x}(w_i)}$. Note that $\eta$ is learning rate, which is constant and can be overlooked. Eq.~\eqref{eq_4} can be regarded as one-step full fine-tuning with batch samples.

\subsubsection{Invertible Transformation}
\label{sec_4_2}

\begin{figure}[t]
  \centering
  \begin{subfigure}[t]{0.45\linewidth}
    \centering
    \includegraphics[width=\linewidth]{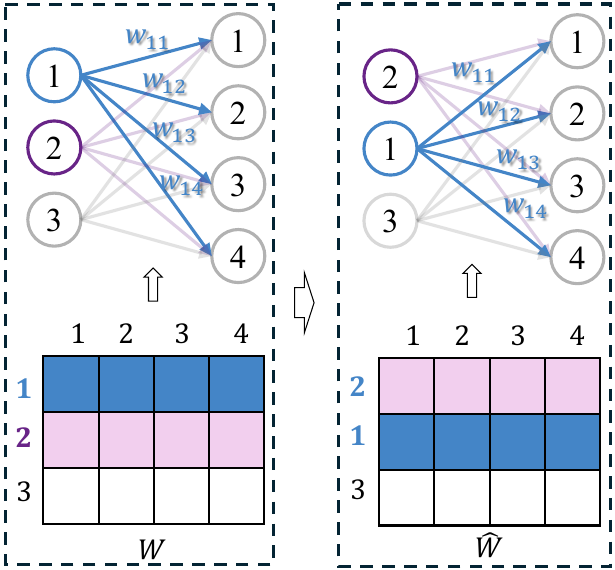}
    \caption{Switch rows 1 and 2 of $W$.}\label{fig3_11}
  \end{subfigure}\hfill
  \begin{subfigure}[t]{0.45\linewidth}
    \centering
    \includegraphics[width=\linewidth]{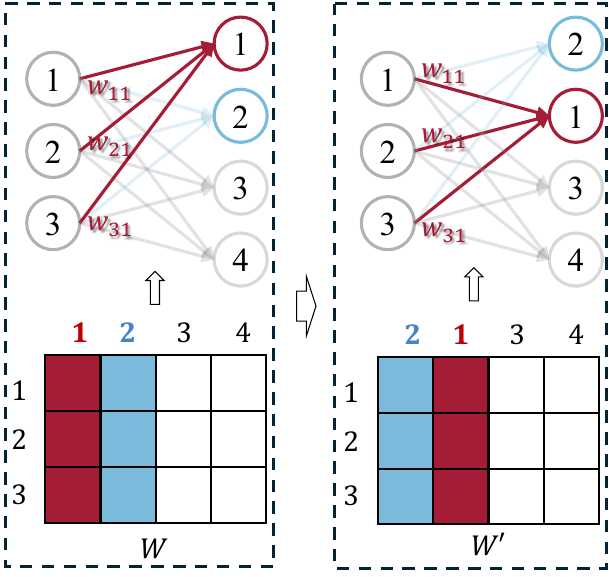}
    \caption{Switch columns 1 and 2 of $W$.}\label{fig3_21}
  \end{subfigure}
  \vspace{-5pt}
  \caption{%
  Rows/columns rearrangement only affects the order of inputs and outputs, without disrupting the link structure information between input and output neurons. E.g., in (a), after switching the first two rows of $W$, the input neuron \textcolor{blue}{\textcircled{\raisebox{-0.9pt}{1}}} moves down and still associated with output neuron \textcolor{gray}{\textcircled{\raisebox{-0.9pt}{2}}} via $w_{12}$.
  }
  \vspace{-15pt}
  \label{fig31}
\end{figure}
The goal of this step is regarding the pre-estimate the weight change $\Delta \hat{W}$ as a guidance to seek an invertible transformation $T(\cdot)$ that can transform a latent spatial-domain matrix $\Delta \bar{W}$ with sparse spectrum to weight change $\Delta W$. 
In general, the spectrum-sparse matrices typically exhibit local smoothness in the spatial domain. Thus, the transformation we are looking for needs to satisfy: \textbf{Point 1}) the weight change should be smoother after using such transformation; \textbf{Point 2}) the structure information of neuron (i.e., link dependence) should be preserved after applying such transformation; and \textbf{Point 3}) the transformation must be invertible since we need to accurately turn to the original weight space to perform PEFT. Fortunately, we find that  the above three points can be well satisfied by the transformation of rows and columns rearrangement: 1) we can rearrange the rows and columns of  weight change to minimizes the difference between adjacent rows or columns, thereby achieving \textbf{Point 1}; 2) as shown in Figure~\ref{fig31}, rearranging the rows and columns of weight matrix does not destroy the link dependence of neurons, which satisfies \textbf{Point 2}; and 3) rows and columns rearrangement is a typical invertible transformation, i.e.,  satisfying \textbf{Point 3}.

Based on this, we formulate the invertible transformation as a solution of row and column rearrangement problem on the pre-estimated weight change $\Delta \hat{W} \in \mathbb{R}^{d \times k}$. The goal is to identify a permutation of rows (or columns) that minimizes the total distance between adjacent rows (or columns), thereby achieving a latent smoother spatial-domain matrix. While alternative approaches may exist, we leave the exploration of such methods to future work. Formally, let $\mathcal{G} = (\mathcal{V}, \mathcal{E})$ be a complete weighted graph, where each node $v_i \in \mathcal{V}$ corresponds to a row (or column) of the matrix $\Delta \hat{W}$, and each edge $e_{ij} \in \mathcal{E}$ connects node $v_i$ and $v_j$ with an associated weight $\beta_{ij}$. The edge weight $\beta_{ij}$ is defined as Euclidean distance between row $i$ and row $j$. We aim to find a permutation $\pi = (\pi_1, \pi_2, \ldots, \pi_n)$ of row (or column) indices that minimizes the total weight of adjacent pairs in the rearranged sequence. That is,
\begin{equation}
\arg\min_{\pi \in \mathcal{P}_n} \sum_{k=1}^{n-1} \beta_{\pi_k, \pi_{k+1}},
\label{eq5}
\end{equation}
where $\mathcal{P}_n$ is the set of all permutations of $\{1, \dots, n\}$ ($n=d$ when rearranging row otherwise $n=k$). %The above problem is akin to a minimum-weight Hamiltonian path problem on the graph $\mathcal{G}$. 

%%%%% lin kenghong change %%%%%%
\textbf{Proposition 1.}
For the rearranged matrix $\Delta \bar{W}$, 
the objective in Eq.~\eqref{eq5} can be equivalently expressed as:
\begin{equation}
\begin{aligned}
    \sum_{k=1}^{n-1} \beta_{\pi_k, \pi_{k+1}}
    &= \sum_{k=1}^{n-1} |\Delta \bar{W}_{k+1}-\Delta \bar{W}_{k}|^2\\
    &= \sum_{u=0}^{n/2-1} 8\sin^2(\frac{\pi u}{n}) |DFT(\Delta \bar{W})_u|^2 ,
\end{aligned}
\end{equation}
where $\Delta \bar{W}_k$ denotes the $k$-th row (or column) of $\Delta \bar{W}$, $DFT(\Delta \bar{W})_u$ denotes the $u$-th row (or column) of the spectral coefficients of $\Delta \bar{W}$. From the proposition, we can see that the objective in Eq.~\eqref{eq5} corresponds to a weighted sum of spectral energies, where each frequency component is weighted by $8\sin^2(\frac{\pi u}{n})$.
The weighting term $8\sin^2(\frac{\pi u}{n})$ grows monotonically with the frequency index $u$ for $u \in [0, n/2-1]$ (for real-valued $\Delta \bar{W}$, the spectrum exhibits conjugate symmetry, so only the first half is considered), meaning that higher frequencies are assigned larger weights.
Therefore, minimizing the left-hand side of Eq.~\eqref{eq5} suppresses the energy of high-frequency components in $\Delta \bar{W}$ and forces most of its spectral energy to concentrate in the low-frequency.
As a result, Eq.~\eqref{eq5} implicitly encourages the rearranged matrix $\Delta \bar{W}$ to exhibit a sparse spectrum.
%%%%% lin kenghong change %%%%%%

\begin{figure}
    \begin{minipage}[t]{1.0\linewidth}
        \begin{algorithm}[H]
        \caption{Nearest Neighbor Search (row rearrangement)}
        \label{alg:sortrc}
        \begin{algorithmic}[1]
            \STATE {\bfseries Input:} Matrix $\Delta \hat{W} \in \mathbb{R}^{d \times k}$; Euclidean distance $\beta_{ij}$
            \STATE {\bfseries Output:} Permutation $\pi = (\pi_1, \pi_2, \ldots, \pi_n)$
            \STATE Initialize set of unvisited indices $\mathcal{U} \leftarrow \{1, 2, \dots, n\}$\;
            \STATE Randomly select a starting index $i \in \mathcal{U}$ and set $\pi_1 \leftarrow i$\;
            \STATE Remove $i$ from $\mathcal{U}$\; 
            \FOR{$k \leftarrow 2$: $n$}
            \STATE Let $j^* = \arg\min_{j \in \mathcal{U}} \beta_{\pi_{k-1}, j}$\;
            \STATE Set $\pi_k \leftarrow j^*$\;
            \STATE Remove $j^*$ from $\mathcal{U}$\;
            \ENDFOR
            % \STATE Compute the sum $S_c \in \mathcal{R}^{1 \times k}$ of $I_W$ along columns
            % \STATE Get sorting index $idx_c$ of $S_c$ along descending order
            % \STATE Reorganize $W$ along $idx_c$ at column: $W = W[:, idx_c]$
            \STATE {\bfseries return} $\pi$
        \end{algorithmic}
        \end{algorithm}% % % % % % % % % % % % % % 
    \end{minipage}
    \vspace{-10pt}
\end{figure}

The Eq.\ref{eq5} is a typical NP-hard problem. In this paper, we adopt a nearest neighbor search algorithm \cite{HURKENS20041} to solve it. As shown in Algorithm\ref{alg:sortrc}, we iteratively select the unvisited node that is most similar to the current node (i.e., the one with the smallest distance $\beta_{ij}$), thereby constructing a permutation in a greedy manner. 
As a result, we can obtain a permutation $\pi^{r}$ (or $\pi^{c}$) of the row (or column) indices, which is used to transform the pre-estimated weight change $\Delta \hat{W}$ into a spatial-domain matrix $\Delta \bar{W}$ with a sparse spectrum. 
It is important to note that this transformation merely involves re-indexing the rows and columns. Therefore, the invertible transformation $T(\cdot)$ in Eq.~\eqref{eq_3}, which maps the spatial-domain matrix $\Delta \bar{W}$ back to the weight change $\Delta W$, can be efficiently implemented via the inverse re-indexing operation, i.e., $\Delta W[\pi_{r}, :][:, \pi_{c}] = \Delta \bar{W}$.

\subsubsection{Spectral Coefficient Sampling}
\label{sec_4_3}
\begin{figure}[t]
  \centering

  \begin{subfigure}[t]{0.19\linewidth}
    \centering
    \includegraphics[width=\linewidth]{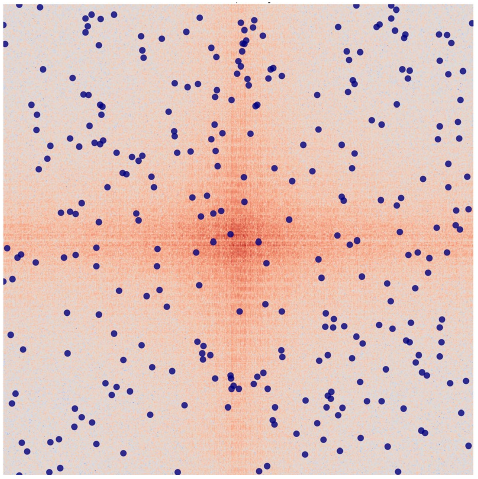}
    \caption{Random}\label{fig3_1}
  \end{subfigure}\hfill
  \begin{subfigure}[t]{0.19\linewidth}
    \centering
    \includegraphics[width=\linewidth]{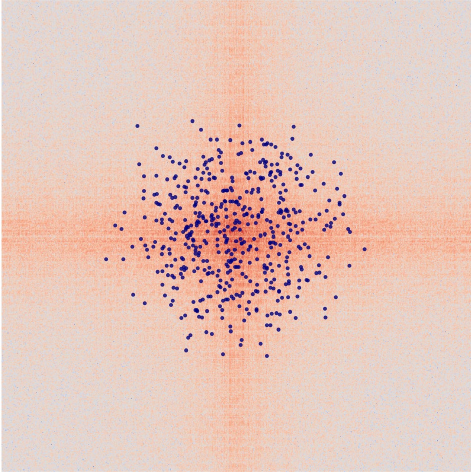}
    \caption{LF}\label{fig3_2}
  \end{subfigure}\hfill
  \begin{subfigure}[t]{0.19\linewidth}
    \centering
    \includegraphics[width=\linewidth]{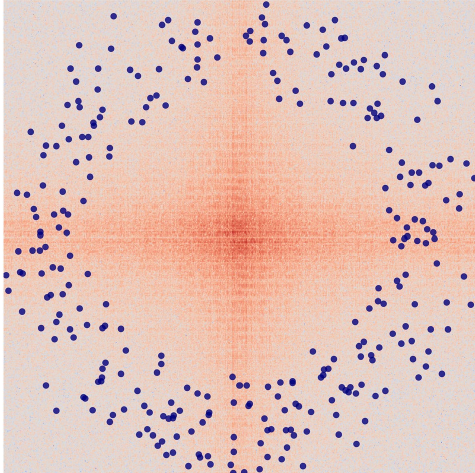}
    \caption{MF}\label{fig3_3}
  \end{subfigure}\hfill
  \begin{subfigure}[t]{0.19\linewidth}
    \centering
    \includegraphics[width=\linewidth]{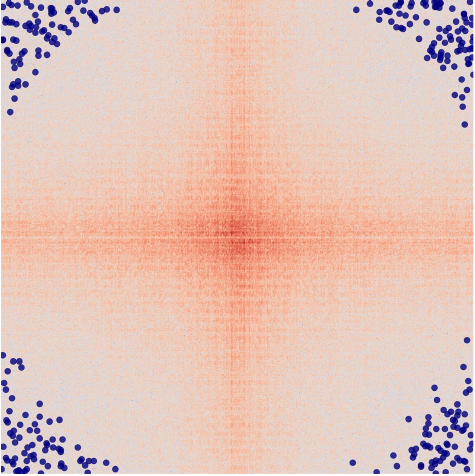}
    \caption{HF}\label{fig3_4}
  \end{subfigure}\hfill
  \begin{subfigure}[t]{0.19\linewidth}
    \centering
    \includegraphics[width=\linewidth]{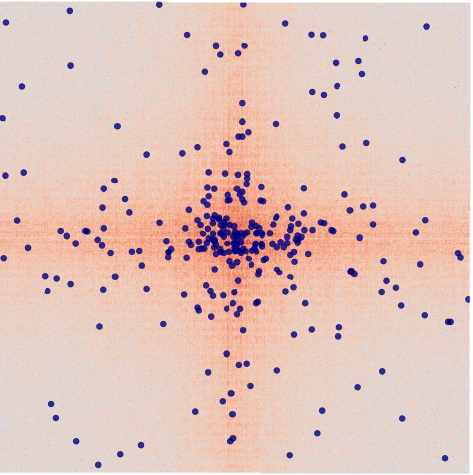}
    \caption{Ours}\label{fig3_5}
  \end{subfigure}
  \vspace{-5pt}
  \caption{Distribution of spectrum and sampling points. LF/MF/HF bias sampling toward low/middle/high frequency.}
  \vspace{-15pt}
  \label{fig3}
\end{figure}
The spectrum of S$^2$FT is sparse, which is different from previous FourierFT \cite{gao2024parameter}. Therefore, a remaining question is should we use the same coefficient sampling strategy? After conduct existing sampling strategy of FourierFT (i.e., randomly sampling or sampling with a bias towards a favored frequency), we find that an interesting result different from FourierFT \cite{gao2024parameter}: \emph{ S$^2$FT performs best when the spectral coefficients are sampled with a bias towards a low frequency, and such strategy is applicable to all evaluated tasks} (see Figure 1 in Appendix). To figure out the reason, we visualize the power spectrum of the latent spatial-domain matrix $\Delta \bar{W}$ and the sampling spectrum points in Figure~\ref{fig3}. It can be seen that 1) our S$^2$FT indeed finds an invertible transformation that can transform the pre-estimated weight change $\Delta \hat{W}$ to a spatial-domain matrix $\Delta \bar{W}$ with sparse spectrum; and 2) among existing sampling strategies, biasing toward low frequencies (LF) performs best, but it still fails to accurately match the spectrum distribution (see Figure~\ref{fig3_2}).

Inspired by this observation, as shown in Figure~\ref{fig3_5}, we propose to regard the power-spectrum of $\Delta \bar{W}$ as a prior to estimate the sampling probability of spectral coefficient: 
\begin{equation}
    p(u,v) = \frac{\|DFT(\Delta \bar{W})\|^{\gamma}_{(u,v)}}{\sum_{u,v} \|DFT(\Delta \bar{W})\|^{\gamma}_{(u,v)}}.
 \label{eq7}
\end{equation}
where $(u, v)$ denotes position, $0 \leq u \leq d$ and $0 \leq v \leq k$; $\|\cdot\|$ denotes amplitude operation; $\gamma$ is a hyper-parameter controlling distribution smoothness ($\gamma=1.5$ is used experientially); $DFT(\cdot)$ denotes discrete Fourier transform. Finally, we sample few trainable spectrum points for complex matrix $F$ of Eq.~\eqref{eq_3} by following the probability $p(u,v)$.

%% file: sec/4_experiment.tex
\begin{table*}[!ht]
    \centering
        \resizebox{1.9\columnwidth}{!}{
        \begin{tabular}{c|ccccccc|cccc|cccccccc|cc}
        \toprule
         \multirow{2}{*}{\diagbox[height=8\line]{Method \\ \\ \\}{\\ \\ \\ Dataset}} & \multicolumn{7}{c}{Natural}                                       & \multicolumn{4}{c}{Specialized}                   & \multicolumn{8}{c}{Structured}                                                                                      & \multicolumn{2}{c}{VTAB}     \\
        \cline{2-22}
         & \rotatebox{90}{CIFAR-100} & \rotatebox{90}{Caltech101} & \rotatebox{90}{DTD} & \rotatebox{90}{Flowers102} & \rotatebox{90}{Pets} & \rotatebox{90}{SVHN} & \rotatebox{90}{Sun397} & \rotatebox{90}{Patch Camelyon} & \rotatebox{90}{EuroSAT} & \rotatebox{90}{Resisc45} & \rotatebox{90}{Retinopathy} & \rotatebox{90}{Clevr/count} & \rotatebox{90}{Clevr/distance} & \rotatebox{90}{DMLab} & \rotatebox{90}{KITTI/distance} & \rotatebox{90}{dSprites/loc} & \rotatebox{90}{dSprites/ori} & \rotatebox{90}{SmallNORB/azi} & \rotatebox{90}{SmallNORB/ele} & \rotatebox{90}{Mean Acc.} & \rotatebox{90}{Mean Params. (\%)} \\
        \midrule
        Full~\cite{jia2022visual}                            & 68.9      & 87.7       & 64.3 & 97.2       & 86.9 & 87.4 & 38.8   & 79.7           & 95.7    & 84.2     & 73.9        & 56.3        & 58.6           & 41.7  & 65.5           & 57.5         & 46.7         & 25.7          & 29.1          & 65.5     & 100.00           \\
        % \midrule
        Linear~\cite{jia2022visual}                          & 63.4      & 85.0       & 64.3 & 97.0       & 86.3 & 36.6 & 51.0   & 78.5           & 87.5    & 68.6     & 74.0        & 34.3        & 30.6           & 33.2  & 55.4           & 12.5         & 20.0         & 9.6           & 19.2          & 53.0     & 0.05             \\
        Bias~\cite{zaken2022bitfit}                            & 72.8      & 87.0       & 59.2 & 97.5       & 85.3 & 59.9 & 51.4   & 78.7           & 91.6    & 72.9     & 69.8        & 61.5        & 55.6           & 32.4  & 55.9           & 66.6         & 40.0         & 15.7          & 25.1          & 62.0     & 0.16             \\
        % \midrule
        Adapter~\cite{houlsby2019parameter}                         & 74.1      & 86.1       & 63.2 & 97.7       & 87.0 & 34.6 & 50.8   & 76.3           & 88.0    & 73.1     & 70.5        & 45.7        & 37.4           & 31.2  & 53.2           & 30.3         & 25.4         & 13.8          & 22.1          & 55.8     & 0.31             \\
        % \bottomrule
         LoRA~\cite{hulora}                         & 68.1      & 91.4      & 69.8 & 99.0       & 90.5 & 86.4 & 53.1   & 85.1           & 95.8    & 84.7     & 74.2        & 83.0        & 66.9           & 50.4  & 81.4           & 80.2         & 46.6         & 32.2          & 41.1          & 72.6    & 0.37            \\
         LoRA-FA~\cite{zhang2023lora}                        & 65.5      & 89.7       & 71.2 & 99.1       & 90.6 & 86.7 & 54.3   & 83.7           & 92.6    & 82.9     & 74.9        & 61.9        & 61.4           & 43.5 & 72.3           & 68.7         & 45.6         & 24.2          & 28.4          &   68.2   & 0.17             \\
         LoRA+~\cite{hayouloraa}                        & 69.6      & 92.5     & 70.8 & 98.5       & 90.4 & 89.8 & 53.8   & 79.9           & 96.1    & 86.5     & 76.5        & 84.4        & 65.1           & 52.8  & 81.6 & 78.6         & 44.2        & 30.5          & 40.3          & 72.7     & 0.37             \\
          HydraLoRA~\cite{tian2024hydralora}                        & 68.8      & 92.0       & 71.3 & 98.8       & 90.4 & 88.0 & 54.6   &84.5           & 96.0    &86.5    & 74.8       & 84.8        & 66.6           & 53.2   & 82.0           & 78.0         & 46.0        & 34.5         & 43.4         & 73.3     & 0.56             \\
          GPS~\cite{zhang2024gradient} & 70.8 & 93.9 & 74.8 & 99.4 & 82.4 & 91.4 & 51.6 &  87.2 & 95.7 & 86.1 & 76.1 &  80.9 & 61.8 & 54.0 & 81.4 & 84.2 & 52.6 & 30.2 & 45.5 & 73.6 & 0.25  \\
        SPT-LoRA-8~\cite{he2023sensitivity} & 72.8 & 92.7 & 72.6 & 99.3 & 91.2 & 86.7 & 55.2 &  85.5 & 95.8 & 85.7 & 75.6 &  82.0 & 68.1 & 49.4 & 81.4 & 80.2 & 48.6 & 29.4 & 39.7 &  73.3 & 0.51             \\

        \midrule
        FourierFT~\cite{gao2024parameter}                       & 67.9     & 89.7     & 72.2 & 99.1       & 91.0 & 90.8 & 55.0   & 85.5           & 95.7    & 86.3     & 75.6       & 82.5        & 68.5           & 52.9  & 81.0 & 75.4         & 46.0        & 28.1          & 38.9          & 72.8     & 0.16            \\
        \rowcolor{mygray} S$^2$FT (Ours)                        & 68.8     & 91.4      & 72.6 & 99.2      & 91.3 & 90.7 & 55.1   & 86.2       & 96.1   & 86.3   & 76       & 82.8       & 69.6           & 52.7 & 81.4       & 79.4       & 46.8        & 29.7          & 41.4         &73.6   &        0.08       \\
        \rowcolor{mygray} S$^2$FT (Ours)                        & 72.0     & 91.9      & 72.5 & 99.2      & 91.5 & 90.6 & 55.7   & 86.9       & 96.2   & 86.7   & 76.5       & 83.8       & 69.7           & 53.9  & 81.7       & 80.1       & 48.4        & 30.5        & 42.0         &\textbf{74.1}    &        0.16       \\
        \bottomrule
        \end{tabular}
        }
    \vspace{-5pt}
    \caption{Results of image classification on VTAB-1k with ViT-B/16 pre-trained on ImageNet-21K.}
    \vspace{-15pt}
    \label{tab:vtab}
\end{table*}

\section{Experiments}
\subsection{Datasets and Evaluation Protocol}
The experiments are conducted on four common tasks. \textbf{1) Image Classification:} VTAB \cite{zhai2019large} and FGVC \cite{wah2011caltech, van2015building, nilsback2008automated, khosla2011novel, gebru2017fine} datasets. The VTAB comprises 19 distinct visual classification tasks organized into 3 semantic domains, we use data splits following \cite{he2023sensitivity}.
The FGVC benchmark includes 5 datasets: CUB-200-2011, NABirds, Oxford Flowers, Stanford Cars, and Stanford Dogs, using standardized data splits from \cite{he2023sensitivity}. Classification accuracy is used as evaluation protocol.
\textbf{2) Image Generation:} the subject-driven text-to-image generation task. We use the dataset proposed in \cite{dreambooth}, where 5 or 6 image samples are used for training each subject. The comparison metrics includes subject fidelity (DINO~\cite{Caron_Touvron_Misra_Jegou_Mairal_Bojanowski_Joulin_2021}, CLIP-I~\cite{Radford_Kim_Hallacy_Ramesh_Goh_Agarwal_Sastry_Amanda_Mishkin_Clark_etal._2021}), text prompt fidelity (CLIP-T~\cite{Radford_Kim_Hallacy_Ramesh_Goh_Agarwal_Sastry_Amanda_Mishkin_Clark_etal._2021}), and sample diversity (LPIPS~\cite{Zhang_Isola_Efros_Shechtman_Wang_2018}). 
\textbf{3) Natural Language Understanding.}
Following \cite{gao2024parameter}, we select 6 tasks (i.e., SST-2, MRPC, CoLA, QNLI, RTE, STSB) from the GLUE (General Language Understanding Evaluation\cite{Wang_Singh_Michael_Hill_Levy_Bowman_2018}) benchmark to compare our S$^2$FT and baselines, where Acc, MCC, and PCC are used as metrics.
\textbf{4) Instruction Tuning.}
For instruction tuning, we conducted experiments on the Alpaca dataset~\cite{wang2023self}. During evaluation, following\cite{gao2024parameter} the fine-tuned models are used to answer a set of standardized questions sourced from the MT-Bench~\cite{zheng2023judging}and Vicuna~\cite{chiang2023vicuna} Eval benchmark suites. The generated responses are then scored by GPT-4 on a scale from 0 to 10.

\begin{table}
\centering
\footnotesize
% \vspace{2pt}
\resizebox{\linewidth}{!}{
\begin{tabular}{c|ccccccc}
\toprule
Dataset & \makecell[c]{CUB\\-2011} & \makecell[c]{NA-\\Brids} & \makecell[c]{Oxford\\Flowers} & \makecell[c]{Stan.\\Dogs} & \makecell[c]{Stan.\\Cars} & \makecell[c]{Mean\\Acc.} & \makecell[c]{Params.\\(\%)} \\
\midrule
Full~\cite{jia2022visual}         & 87.3 & 82.7 & 98.8 & 89.4 & 84.5 & 88.5 & 100.00 \\
Linear~\cite{jia2022visual}       & 85.3 & 75.9 & 97.9 & 86.2 & 51.3 & 79.3 & 0.21 \\
Bias~\cite{zaken2022bitfit}       & 88.4 & 84.2 & 98.8 & 91.2 & 79.4 & 88.4 & 0.33 \\
Adapter~\cite{houlsby2019parameter} & 87.1 & 84.3 & 98.5 & 89.8 & 68.6 & 85.6 & 0.48 \\
LoRA~\cite{hulora}                & 84.9 & 79.0 & 98.1 & 88.1 & 87.4 & 87.5 & 0.34 \\
LoRA-FA~\cite{zhang2023lora}      & 85.8 & 79.1 & 99.1 & 88.1 & 80.1 & 86.4 & 0.17 \\
LoRA+~\cite{hayouloraa}           & 86.5 & 80.8 & 99.2 & 87.4 & 88.7 & 88.5 & 0.34 \\
HydraLoRA~\cite{tian2024hydralora} & 86.9 & 82.1 & 99.1 & 88.2 & 89.1 & 89.0 & 0.47 \\
\midrule
FourierFT~\cite{gao2024parameter} & 85.5 & 82.4 & 99.1 & 89.0 & 83.7 & 87.9 & 0.16 \\
\rowcolor{mygray} S$^2$FT (Ours) & 87.9 & 84.3 & 99.1 & 90.6 & 83.6 & 89.1 & 0.08 \\
\rowcolor{mygray} S$^2$FT (Ours) & 88.4 & 84.6 & 99.3 & 91.4 & 84.5 & \textbf{89.6} & 0.16 \\
\bottomrule
\end{tabular}
}
\vspace{-5pt}
\captionof{table}{Image classification on FGVC}
\vspace{-20pt}
\label{tab:fgvc}
\end{table}%

\begin{table}

\footnotesize
\centering
\renewcommand{\arraystretch}{1.4}
    \resizebox{\linewidth}{!}{
    \smallskip\begin{tabular}{c|ccccc}
        \hline
        \multicolumn{1}{c|}{\multirow{1}{*}{Method}} & DINO $\uparrow$ & CLIP-I $\uparrow$ & CLIP-T $\uparrow$ & LPIPS $\uparrow$ & Params(\%)\\
        \hline
        RealImages & 0.703 & 0.864  & \textminus & 0.695 & \textminus\\
        % \hline
        DreamBooth & 0.614 & 0.778  & 0.239 & 0.737 & 100\\
         % \hline
        LoRA & 0.613 & 0.765  & 0.237 & 0.744 & 1.44\\
        \hline
        FourierFT &0.607 & 0.750  & 0.237 & 0.732 & 0.07\\
        % \hline
        \rowcolor{mygray} S$^2$FT(Ours) &\textbf{0.620} & \textbf{0.784}  & 0.234 & \textbf{0.769} & 0.06\\
        \hline
\end{tabular}
}
\vspace{-5pt}
\caption{Results of subjective generation task.}\smallskip
\vspace{-20pt}
\label{table44}
\end{table}

\subsection{Implementation Details and Baselines}
\noindent \textbf{Image Classification.} We follow \cite{he2023sensitivity} to process the images of FGVC and VTAB-1k. We employ the AdamW optimizer with cosine learning rate decay to fine-tune models for 100 epochs and 10 epochs linear warm-up. The baseline methods include reparameterization-based methods such as LoRA \cite{hulora}, (LoRA-FA\cite{zhang2023lora}, LoRA+\cite{hayouloraa}, HydraLoRA\cite{tian2024hydralora}), FourierFT\cite{gao2024parameter} and other latest methods. In particular, we sample 6000 points for FourierFT and 3000 points for our S$^2$FT for parameter efficiency. 

\noindent \textbf{Image Generation.} We set hyperparameter by following \cite{dreambooth} and select latest Dreambooth, LoRA and FourierFT as our baseline. Here, we sample 10000 and 8000 points for FourierFT and our S$^2$FT.  

\noindent \textbf{Natural Language Understanding.} Following \cite{gao2024parameter}, we compare our S$^2$FT with wide range of models, including Adapter, LoRA, AdaLoRA, DyLoRA, CorDA, FourierFT. The methods are fine-tuned on the commonly used RoBERTa model, including both base and large version. 

\noindent \textbf{Instruction Tuning.} Following \cite{gao2024parameter}, we apply LoRA, FourierFT and our S$^2$FT to fine-tune several base models from the LLaMA2 family, including LLaMA2-7B and LLaMA2-13B. Unless otherwise noted, we use the same hyperparameter settings as FourierFT for fair comparison.

% \noindent \textbf{Instruction Tuning.} Following \cite{gao2024parameter}, we apply LoRA, FourierFT and our S$^2$FT to fine-tune several base models from LLaMA2 family, including LLaMA2-7B and LLaMA2-13B. These models are fine-tuned on the Alpaca dataset, which contains about 51,000 high-quality instruction-following samples generated by text-davinci-003 (GPT-3.5). During the evaluation phase, we use the fine-tuned models to answer a set of standardized questions, which are sourced from the MT-Bench and Vicuna Eval benchmark suites. The generated responses are then scored by GPT-4 on a scale from 0 to 10. To ensure experimental fairness, we use the same hyperparameter configuration as employed in FourierFT.

\begin{table}
\centering
\footnotesize
\resizebox{\linewidth}{!}{
\begin{tabular}{@{}l|r|ccccccc@{}}
\toprule
\textbf{Model} & \textbf{Params} & SST-2 & MRPC & CoLA & QNLI & RTE & STS-B & Avg. \\
 & (\%) & (Acc.) & (Acc.) & (MCC) & (Acc.) & (Acc.) & (PCC) & \\
\midrule
$\rm{RoB_{base}}$(Full) & 100.0 & 94.8 & 90.2 & 63.6 & 92.8 & 78.7 & 91.2 & 85.2 \\
BitFit & 0.08 & 93.7 & \textbf{92.7} & 62 & 91.8 & \textbf{81.5} & 90.8 & 85.4 \\
Adapter$^{\text{D}}$ & 0.24 & 94.2 & 88.5 & 60.8 & 93.1 & 71.5 & 89.7 & 83.0 \\
Adapter$^{\text{D}}$ & 0.72 & 94.7 & 88.4 & 62.6 & 93.0 & 75.9 & 90.3 & 84.2 \\
LoRA & 0.24 & 95.1 & 89.7 & 63.4 & \textbf{93.3} & 78.4 & \textbf{91.5} & 85.2 \\
AdaLoRA & 0.24 & 94.5 & 88.7 & 62.0 & 93.1 & 81.0 & 90.5 & 85.0 \\
DyLoRA & 0.24 & 94.3 & 89.5 & 61.1 & 92.2 & 78.7 & 91.1 & 84.5 \\
CorDA & 0.24 & 93.1 & 89.7 & 59.6 & 91.5 & 76.2 & 90.2 & 83.4 \\
FourierFT & 0.02 & 94.2 & 90.0 & 63.8 & 92.2 & 79.1 & 90.8 & 85.0 \\
\rowcolor{mygray} S$^2$FT & 0.02 & \textbf{95.3} & 90.4 & \textbf{64.8} & 92.8 & 79.8 & 90.7 & \textbf{85.6} \\
\midrule
$\rm{RoB_{large}}$(Full) & 100.0 & 96.4 & \textbf{90.9} & 68.0 & 94.7 & 86.6 & \textbf{92.4} & 88.2 \\
Adapter$^{\text{P}}$ & 0.86 & 96.1 & 90.2 & 68.3 & \textbf{94.8} & 83.8 & 92.1 & 87.6 \\
Adapter$^{\text{P}}$ & 0.22 & \textbf{96.6} & 89.7 & 67.8 & \textbf{94.8} & 80.1 & 91.9 & 86.8 \\
% Adapter$^{\text{H}}$ & 1.65 & 96.2 & 88.7 & 66.5 & 94.7 & 83.4 & 91.0 & 86.8 \\
% Adapter$^{\text{H}}$ & 0.22 & 96.3 & 87.7 & 66.3 & 94.7 & 72.9 & 91.5 & 84.9 \\
LoRA & 0.22 & 96.2 & 90.2 & 68.2 & \textbf{94.8} & 85.2 & 92.3 & 87.8 \\
DyLoRA & 0.22 & 94.7 & 90.7 & 65.3 & 93.6 & 87.2 & 91.4 & 87.2 \\
PiSSA & 0.22 & 95.8 & 91.5 & 68.1 & 94.4 & 87.9 & 92.0 & 88.2 \\
FourierFT & 0.01 & 96.0 & \textbf{90.9} & 67.1 & 94.4 & 87.4 & 91.9 & 88.0 \\
\rowcolor{mygray} S$^2$FT & 0.01 & 96.4 & 90.7 & 67.6 & 94.6 & \textbf{88.1} & 92.3 & \textbf{88.3} \\
\bottomrule
\end{tabular}
}
\vspace{-5pt}
\captionof{table}{GLUE results with RoBERTa}
\vspace{-5pt}
\label{tab:glue}
% \end{minipage}
\end{table}

\begin{table}[!ht]
\scriptsize 
\centering
\resizebox{1\linewidth}{!}{
\begin{tabular}{@{}l|l|r|cc@{}}
\toprule
Model & Method & \begin{tabular}[c]{@{}r@{}}\# Trainable\\ Parameters\end{tabular} & MT-Bench & Vicuna \\ 
\midrule

\multirow{4}{*}{LLaMA2-7B} 
& LoRA\dag          & 159.9M  & 5.19\textsubscript{$\pm$.1} & 7.38\textsubscript{$\pm$.3} \\ 
& LoRA              & 33.5M   & 5.20\textsubscript{$\pm$.3} & 7.35\textsubscript{$\pm$.6} \\ 
& FourierFT         & 0.064M  & 5.18\textsubscript{$\pm$.3} & 7.49\textsubscript{$\pm$.4} \\ 
& \textbf{S$^2$FT}  & 0.064M  & \textbf{5.21}\textsubscript{$\pm$.4} & \textbf{7.50}\textsubscript{$\pm$.6} \\

\midrule

\multirow{4}{*}{LLaMA2-13B} 
& LoRA\dag          & 250.3M  & 5.78\textsubscript{$\pm$.2} & 7.89\textsubscript{$\pm$.5} \\
& LoRA              & 52.4M   & 5.80\textsubscript{$\pm$.2} & 7.89\textsubscript{$\pm$.6} \\
& FourierFT         & 0.08M   & 5.82\textsubscript{$\pm$.3} & 7.92\textsubscript{$\pm$.5} \\
& \textbf{S$^2$FT}  & 0.08M   & \textbf{5.89}\textsubscript{$\pm$.4} & \textbf{7.94}\textsubscript{$\pm$.6} \\

\bottomrule
\end{tabular}
}
\vspace{-5pt}
\caption{The average scores on MT-Bench and Vicuna assessed by GPT-4. $\dag$ indicates updating the layers other than \texttt{lm\_head}. Higher score is better.}
\vspace{-10pt}
\label{tab:instruction}
\end{table}

\subsection{Discussion of Results}
\noindent \textbf{Image Classification Tasks.} In Tables~\ref{tab:vtab} and \ref{tab:fgvc}, we report the performance of our S$^2$FT and baselines on VTAB and FGVC. It can be seen that 1) compared with recent PEFT methods, our S$^2$FT achieves superior performance on most datasets, which verifies the effectiveness of our S$^2$FT; 2) In Fourier transforms-based PEFT method, our S$^2$FT method achieves a significant improvement (around 1\% $\sim$ 2\%). In particular, our method uses only 50\% of the parameters compared to FourierFT. This indicates that our method better leverages the expressive power of the spectrum domain.

\noindent \textbf{Image Generation Tasks.} From Table~\ref{table44}, we can see that (1)  S$^2$FT significantly outperforms FourierFT on CLIP-I and LPIPS, while achieving slightly better or comparable performance on DINO and CLIP-T; and (2) our method achieves performance comparable to DreamBooth (100\%) and LoRA (1.44\%) with only 0.06\% of the total parameters. The results suggest that our method not only reduces the number of trainable parameters but also yields better generation quality. %Beside, from qualitative results of Figure~\ref{fig4}, we also can observe that  S$^2$FT achieves better object  consistency while keeping a harmonious  image style . %For example, the color of berry in the bow keeps the same deep purple as the reference image, while the FourierFT generates some orange berries. Besides, the image of cat and dog using FourierFT exhibit color distortion issues occasionally, while S$^2$FT is capable of preserving correct color fidelity.

\noindent \textbf{Nature Language Understanding Tasks.} Table~\ref{tab:glue} presents the results of natural language understanding. 
It can be found that our S$^2$FT achieves superior performance on both the base and large versions of RoBERTa in average with the least trainable parameters scale. Compared with FourierFT, our S$^2$FT constantly improves 0.6\% and 0.3\% in average score using the same scale of parameters. Compared with the commonly used LoRA, our S$^2$FT achieves better result using 12 times smaller parameters.  It is worth noting that our S$^2$FT is the only method that even outperforms full fine-tuning with $100\%$ parameters in average, demonstrating the versatility of our approach.

\noindent \textbf{Instruction Tuning Tasks.} Table~\ref{tab:instruction} presents the results of the instruction tuning task. As shown, our S$^2$FT consistently outperforms other methods on both LLaMA2-7B and LLaMA2-13B in average performance, while requiring only a minimal number of trainable parameters. In particular, compared to the widely adopted LoRA, S$^2$FT achieves better performance using merely 0.15\% of its parameter count.
% \noindent \textbf{Can our S$^2$FT method adapt to large language models?} In Table~\ref{tab:glue}, we fine-tune the pre-trained RoBERTa Base and Large foundation models\cite{Liu_Ott_Goyal_Du_Joshi_Chen_Levy_Lewis_Zettlemoyer_Stoyanov_2019} and evaluate them on the GLUE benchmark(General Language Understanding Evaluation\cite{Wang_Singh_Michael_Hill_Levy_Bowman_2018}). At the same time, we compare our method with the main competitor, FourierFT. For fairness, we reuse the results reported in their paper. From the results, it is evident that our method achieves better or comparable performance, demonstrating the versatility of our approach.

\begin{figure}[t]
  \centering

  \begin{subfigure}[t]{0.45\linewidth}
    \centering
    \includegraphics[width=0.9\linewidth]{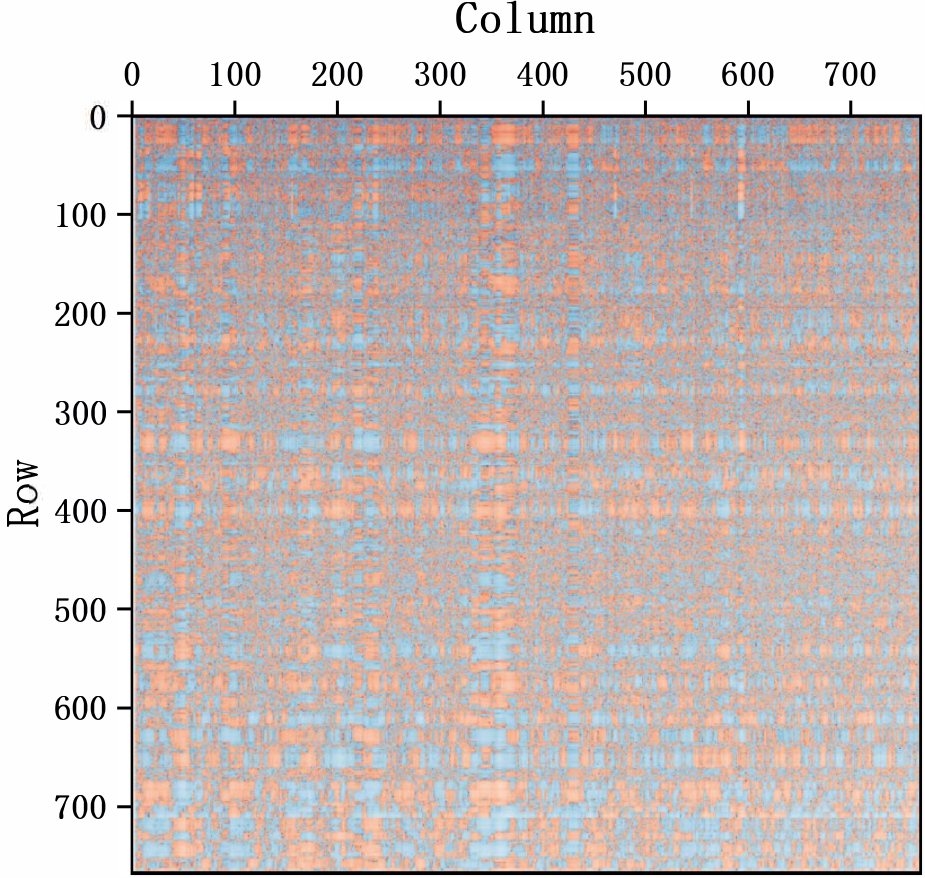}
    \caption{Latent matrix $\Delta \bar{W}$.}\label{fig6_a}
  \end{subfigure}\hfill
  \begin{subfigure}[t]{0.45\linewidth}
    \centering
    \includegraphics[width=0.9\linewidth]{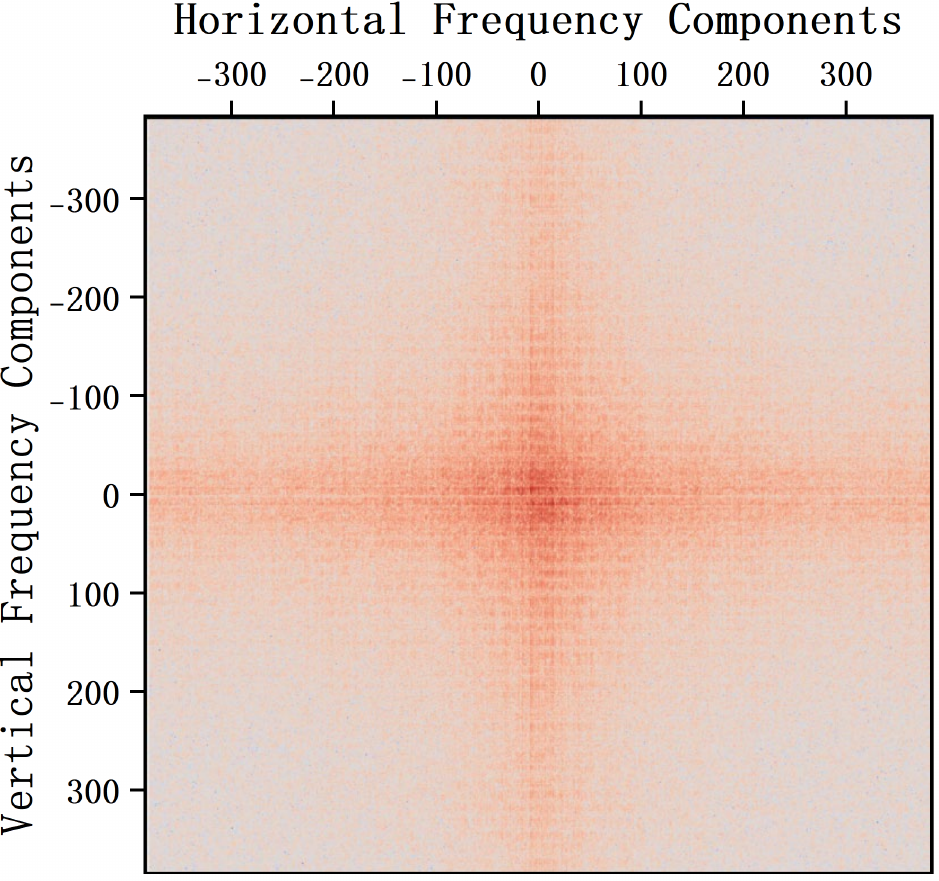}
    \caption{Power spectrum of $\Delta \bar{W}$.}\label{fig6_b}
  \end{subfigure}

  \par\vspace{0.6em}

  \begin{subfigure}[t]{0.45\linewidth}
    \centering
    \includegraphics[width=0.9\linewidth]{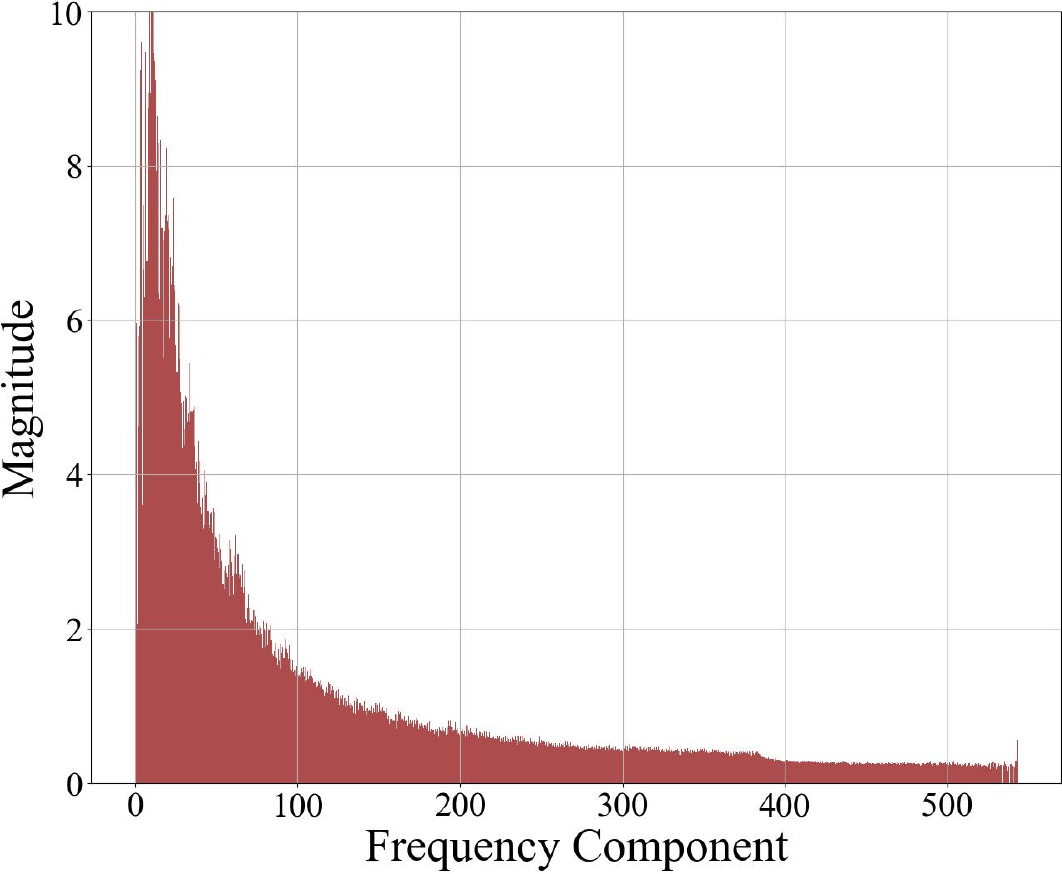}
    \caption{Amplitude distribution.}\label{fig6_c}
  \end{subfigure}\hfill
  \begin{subfigure}[t]{0.45\linewidth}
    \centering
    \includegraphics[width=0.9\linewidth]{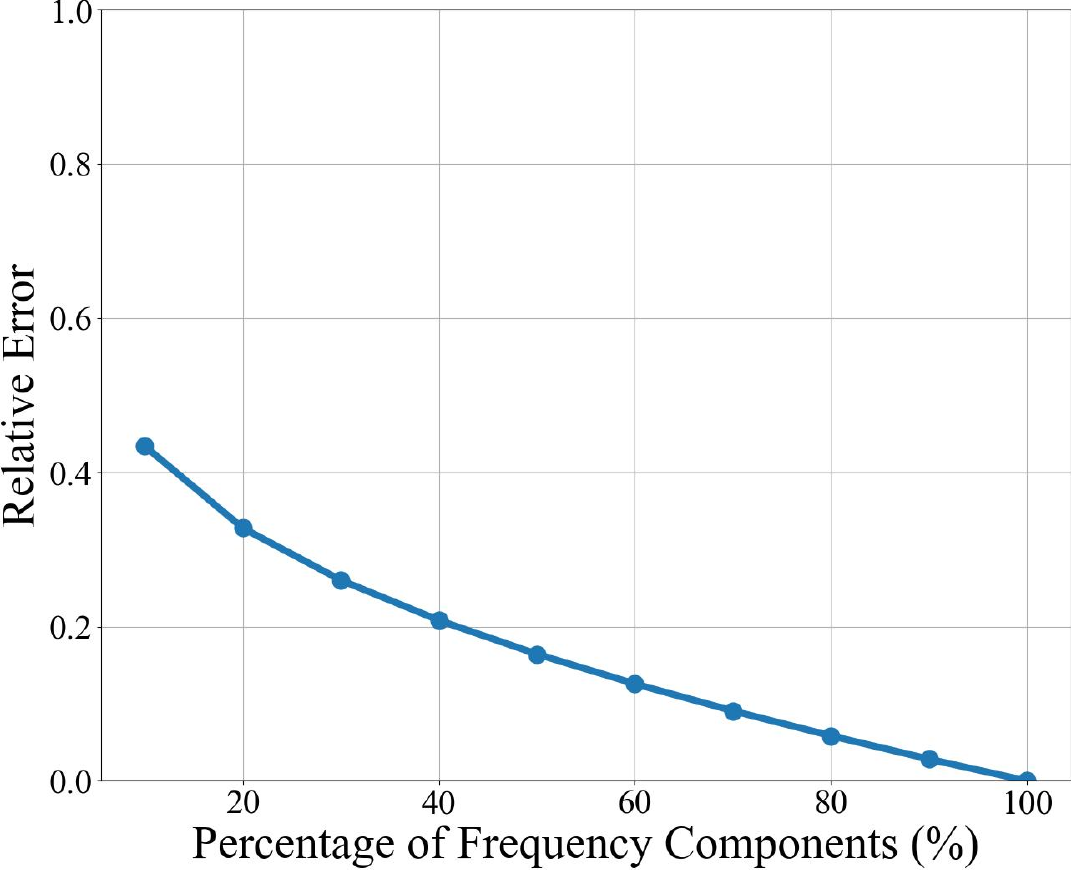}
    \caption{Relative error of $\Delta W$.}\label{fig6_d}
  \end{subfigure}

  \caption{Distribution analysis of the spatial-domain matrix $\Delta \bar{W}$ obtained by our S$^2$FT.}
  \vspace{-15pt}
  \label{fig6}
\end{figure}

\begin{figure*}[!ht]
  \centering
  \includegraphics[width=0.95\linewidth]{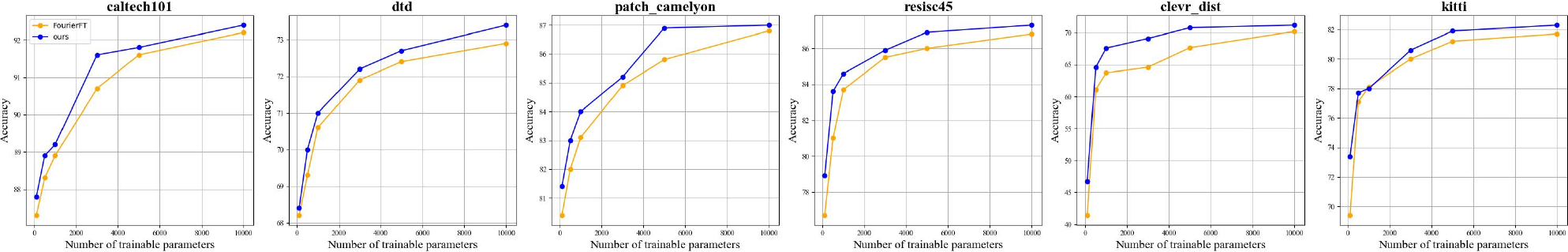}
  \vspace{-5pt}
  \caption{Performance comparison between FourierFT and our S$^2$FT on trainable parameters.}
\label{fig_params}
  \vspace{-10pt}
\end{figure*}

\subsection{Ablation Study}
% \noindent \textbf{How does our S$^2$FT perform under different training parameters?} 
\noindent \textbf{Impact of number of trainable parameters.} 
Figure~\ref{fig_params} reports the accuracy of our S$^2$FT and baseline on eight tasks from the VTAB dataset using different number of trainable parameters.  
It can be found that (1) the accuracy of both methods improves as the number of trainable parameters increases; and (2) our S$^2$FT consistently outperforms baseline, indicating the superiority and efficiency of our S$^2$FT.

\noindent \textbf{Can our S$^2$FT seek a latent spatial-domain matrix $\Delta \bar{W}$ with sparse spectrum?} 
To answer this question, in Figure~\ref{fig6}, we visualize the spatial-domain matrix $\Delta \bar{W}$ obtained by our invertible transformation $T^{-1}()$, and its power spectrum, amplitude distribution, and relative error of modeling $\Delta W$. Compared with FourierFT (see Figure~\ref{fig2}), we find that the spatial-domain matrix $\Delta \bar{W}$ indeed presents smooth property (see Figure~\ref{fig6_a}) and sparse spectrum (see Figures~\ref{fig6_b} and \ref{fig6_c}), and show superior modeling performance with only a few of spectrum coefficients (see Figure~\ref{fig6_d}) . This verifies the superiority of our S$^2$FT on modeling the weight change $\Delta W$ with few coefficients.

\noindent \textbf{How does the accuracy of weight change pre-estimation affect performance?} 
In Table~\ref{tab_pre_method}, we evaluate our S$^2$FT using different pre-estimation methods, including our method defined in Eq.~\ref{eq_4} and the full finetuning (FF) pre-estimation method with different epochs. Note that FF with more optimization epochs yields more accurate pre-estimations, as it better fits the training samples of downstream tasks.
Despite this, the results show that the accuracy of the weight change pre-estimation has little impact on the performance. This is reasonable, as the pre-estimation is only used to guide us seeking a coarse row/column rearrangement, and the method does not rely heavily on its precision.
\begin{table}
\centering
\small
\resizebox{1.0\linewidth}{!}{
	\begin{tabular}{l|c|c|c|c|c}
		\toprule
		Method  & Our Eq.~\ref{eq_4} & FF (1 epoch) & FF (2 epoch)  & FF (3 epoch) & FF (5 epoch) \\ 
		\midrule
        S$^2$FT &  73.6 &  73.6 & 73.7 & 73.7 & 73.8 \\
        \bottomrule
	\end{tabular}}
\vspace{-5pt}
\caption{Impact Analysis of pre-estimation accuracy.}
\vspace{-10pt}
\label{tab_pre_method}
\end{table}
%Specifically, using 20\% training data for pre-estimation achieves the highest accuracy, indicating that we only using few training data with low computational cost is enough to achieve good result. Thus, 20\% data is used in this paper.
% \begin{wraptable}{r}{0.6\linewidth}

% \end{wraptable}
% \noindent \textbf{Is there a frequency bias in our S$^2$FT?} 
% % 倒过来写也可以，直接说我们可以解释，然后开始说实验结果来支持
% To analyze frequency bias, following \cite{gao2024parameter}, we also conduct an experiment using a sampling strategy biased toward different central frequencies to select spectrum points. As shown in Figure~\ref{fig_fc}, our S$^2$FT achieves better performance when favoring lower central frequencies, indicating that the spectrum domain in our method is sparse. Moreover, we observe an interesting but previously unexplained phenomenon noted in the original FourierFT paper \cite{gao2024parameter}: the optimal central frequency for FourierFT varies randomly across tasks, and low frequencies are often not the most effective. This can be attributed to the fact that the spectrum domain in FourierFT is not sparse but rather power-uniform.
\noindent \textbf{Is there a frequency bias in our S$^2$FT?}
In Figure 7 of appendix, we conduct an experiment using a sampling strategy biased toward different central frequencies to select spectrum points. We can see that our S$^2$FT achieves better performance when favoring lower central frequencies, indicating that the spectrum domain in our method is sparse. Moreover, we observe an interesting but previously unexplained phenomenon noted in the original FourierFT paper \cite{gao2024parameter}: the optimal central frequency for FourierFT varies randomly across tasks, and low frequencies are often not the most effective. This can be attributed to the fact that the spectrum domain in FourierFT is not sparse but rather power-uniform.

%To analysis the frequency bias, following \cite{gao2024parameter}, we conduct an experiment by using sampling bias strategy toward different central frequency to sample spectrum points. As shown in Figure~\ref{fig_fc}, our S$^2$FT achieves better performance while using a lower favored central frequency, which qualifies that our spectrum domain is sparse. Besides, we also notice that an interesting but not-well-explained phenomenon, in the origin FourierFT paper \cite{gao2024parameter}, that FourierFT's best central frequency is random for different task, and the low frequency is not the best for FourierFT most of time. Actually, it is because the spectrum domain of FourierFT is not sparse but power-uniform.

% \begin{wraptable}{r}{0.6\linewidth}
\begin{table}
\centering
\small
\resizebox{1.0\linewidth}{!}{
	\begin{tabular}{l|c|c|c|c}
		\toprule
		Method  & Natural & Specialized & Structured & Mean Acc \\ 
		\midrule
        S$^2$FT + Random sampling &  79.6 & 84.9 &58.9 & 72.0 \\
        S$^2$FT + Sampling toward LF & 80.6 & 85.5 &59.7 & 72.8 \\
        \rowcolor{mygray}S$^2$FT + Our sampling &  81.3 & 86.2 &60.5 & \textbf{73.6} \\
        \bottomrule
	\end{tabular}}
\vspace{-5pt}
\caption{Analysis of sampling method (LF is low frequency).}
\vspace{-10pt}
\label{tab_sampling_method}

\end{table}
% \end{wraptable}

\noindent \textbf{Is our spectral coefficient sampling effective?} 
In Table~\ref{tab_sampling_method}, we report the accuracy of our S$^2$FT with different spectrum coefficient sampling strategy on VTAB benchmark. We can see that our proposed sampling strategy consistently performs better with the ones proposed in \cite{brandwood2012fourier}, which verifies the superiority of our sampling strategy.
\begin{table}
\centering
\small
\resizebox{1.0\linewidth}{!}{
\begin{tabular}{c|c|c|c|c}
		\toprule
	Method & Training time / epochs(s) & Trans. / Trans.$^{-1}$(s)&Training Memory (GB) & Params. (\%)\\
	\midrule
    Full &  6.2 & - &18.8 & 100 \\
    LoRA  &  3.8 & - &9.7 & 0.37\\
    FourierFT &  4.0 & - &8.7 & 0.16\\
    S$^{2}$FT &  4.2 & 0.02 &8.5 & 0.08\\
    \bottomrule
\end{tabular}}
\vspace{-5pt}
\caption{Analysis of training cost on the VTAB dataset.}
\vspace{-15pt}
\label{tab:Ablation5}
\end{table}
% \end{wraptable}

\noindent \textbf{Is our S$^2$FT computation efficient?}
In Table~\ref{tab:Ablation5}, we report the training cost comparison of our S$^2$FT, FourierFT, and LoRA using the ViT base model. While consuming similar inference time and memory. Compared with LoRA and FourierFT, S$^2$FT also requires smaller memory cost because our S$^2$FT only consumes smaller training parameter scale.

\noindent \textbf{How does the hyper-parameter $\gamma$ impact the performance of our S$^2$FT?} 
 In Table~\ref{table_9} we report the performance of our S$^2$FT under different values of the hyper-parameter $\gamma$ on VTAB benchmark, which controls the smoothness of sampling probability. As shown, when $\gamma$ increases from 0.5 to 1.5, the mean accuracy steadily improves from 72.3\% to 73.6\%, indicating that a moderate enhancement of smoothness helps better exploit the sparse spectral distribution. The performance peaks at $\gamma=1.5$, where S$^2$FT achieves the best performance across the three categories (Natural, Specialized, Structured). However, further increasing $\gamma$ beyond 1.5 leads to a gradual decline in performance, possibly due to over-smoothing, which may be inconsistent with the underlying distribution information. Therefore, $\gamma=1.5$ is adopted as the default setting in our experiments for its empirical effectiveness.

\begin{table}[!t]
% \vspace{-10pt}
\centering
% \small
\resizebox{1.0\linewidth}{!}{
\begin{tabular}{l|c|c|c|c|c}
\toprule
Method & Params. (\%) & Natural & Specialized & Structured & Mean Acc \\
\midrule
S$^2$FT($\gamma=0.5$) & 0.08 & 80.7 & 84.9 & 58.6 & 72.3 \\
S$^2$FT($\gamma=1.0$)  & 0.08 & 81.1 & 84.8 & 58.9 & 72.5 \\
\rowcolor{mygray} S$^2$FT($\gamma=1.5$)  & 0.08 & 81.3 & 86.2 & 60.5 & 73.6 \\
S$^2$FT($\gamma=2.0$)  & 0.08 & 81.0 & 85.9 & 60.5 & 73.4 \\
S$^2$FT($\gamma=3.0$)  & 0.08 & 80.8 & 85.4 & 60.2 & 73.1 \\
S$^2$FT($\gamma=5.0$)  & 0.08 & 80.0 & 85.1 & 60.1 & 72.7 \\
% \rowcolor{mygray} S$^2$FT(20\%) & 0.08 & 81.1 & 85.9 & 60.8 & 73.6 \\
% S$^2$FT(10\%)  & 0.08 & 81.2 & 85.7 & 60.7 & 73.5 \\
\bottomrule
\end{tabular}
}
\vspace{-5pt}
\caption{Impact analysis on hyper-parameter $\lambda$.}
\vspace{-10pt}
\label{table_9}
\end{table}

%% file: sec/5_conclusion.tex
\section{Conclusions}
In this paper, we identified a key challenge overlooked in Fourier-based parameter-efficient fine-tuning (PEFT), i.e., the power spectrum of weight change $\Delta W$ is not sparse, but tends to uniform. This resulted in that using few spectral coefficients is difficult to accurately model $\Delta W$.
To address this challenge, we presented a novel PEFT method with sparse spectrum domain, which aims to seek an invertible transformation that  transforms a latent spatial-domain matrix with sparse spectrum to the weight change, and then perform PEFT on such sparse spectrum domain with few spectral coefficients. Results showed that our S$^2$FT achieves superior performance over previous methods. 

\section*{Acknowledgments}
This work was supported by the NSFC under Grant No. 62502120, 62272130 and 62501408, the Guangdong Basic and Applied Basic Research Foundation under Grant No.
2025A1515011674, Shenzhen Science and Technology Program No. ZDCYKCX20250901092700001, SYSPG 20241211173609009, and JCYJ20240813142104006.

%% file: main.bib
@String(CVPR= {IEEE Conf. Comput. Vis. Pattern Recog.})

@String(ICCV= {Int. Conf. Comput. Vis.})

@String(ICLR = {Int. Conf. Learn. Represent.})

@String(AAAI = {AAAI})

@String(CVPR  = {CVPR})

@String(ICCV  = {ICCV})

@String(ICLR  = {ICLR})

@InProceedings{fft_1,
  author = {Xu, Kai and Qin, Minghai and Sun, Fei and Wang, Yuhao and Chen, Yen-Kuang and Ren, Fengbo},
  title = {Learning in the Frequency Domain},
  booktitle = {Proceedings of the IEEE/CVF Conference on Computer Vision and Pattern Recognition (CVPR)},
  month = {June},
  year = {2020}
}

@inproceedings{fft_2,
  title={Deep residual learning in the jpeg transform domain},
  author={Ehrlich, Max and Davis, Larry S},
  booktitle={Proceedings of the IEEE/CVF international conference on computer vision},
  pages={3484--3493},
  year={2019}
}

@article{fft_3,
  title={Faster neural networks straight from jpeg},
  author={Gueguen, Lionel and Sergeev, Alex and Kadlec, Ben and Liu, Rosanne and Yosinski, Jason},
  journal={Advances in Neural Information Processing Systems},
  volume={31},
  year={2018}
}

@inproceedings{fft_4,
  title={Rethinking graph neural networks for anomaly detection},
  author={Tang, Jianheng and Li, Jiajin and Gao, Ziqi and Li, Jia},
  booktitle={International conference on machine learning},
  pages={21076--21089},
  year={2022},
  organization={PMLR}
}

@inproceedings{jia2022visual,
  title={Visual prompt tuning},
  author={Jia, Menglin and Tang, Luming and Chen, Bor-Chun and Cardie, Claire and Belongie, Serge and Hariharan, Bharath and Lim, Ser-Nam},
  booktitle={European Conference on Computer Vision},
  pages={709--727},
  year={2022},
  organization={Springer}
}

@inproceedings{zaken2022bitfit,
  title={BitFit: Simple Parameter-efficient Fine-tuning for Transformer-based Masked Language-models},
  author={Zaken, Elad Ben and Goldberg, Yoav and Ravfogel, Shauli},
  booktitle={Proceedings of the 60th Annual Meeting of the Association for Computational Linguistics (Volume 2: Short Papers)},
  pages={1--9},
  year={2022}
}

@inproceedings{houlsby2019parameter,
  title={Parameter-efficient transfer learning for NLP},
  author={Houlsby, Neil and Giurgiu, Andrei and Jastrzebski, Stanislaw and Morrone, Bruna and De Laroussilhe, Quentin and Gesmundo, Andrea and Attariyan, Mona and Gelly, Sylvain},
  booktitle={International conference on machine learning},
  pages={2790--2799},
  year={2019},
  organization={PMLR}
}

@article{hulora,
  title={Lora: Low-rank adaptation of large language models.},
  author={Hu, Edward J and Shen, Yelong and Wallis, Phillip and Allen-Zhu, Zeyuan and Li, Yuanzhi and Wang, Shean and Wang, Lu and Chen, Weizhu and others},
  journal={ICLR},
  volume={1},
  number={2},
  pages={3},
  year={2022}
}

@article{gao2024parameter,
  title={Parameter-Efficient Fine-Tuning with Discrete Fourier Transform},
  author={Gao, Ziqi and Wang, Qichao and Chen, Aochuan and Liu, Zijing and Wu, Bingzhe and Chen, Liang and Li, Jia},
  journal={arXiv preprint arXiv:2405.03003},
  year={2024}
}

@inproceedings{hayouloraa,
  title={LoRA+: Efficient Low Rank Adaptation of Large Models},
  author={Hayou, Soufiane and Ghosh, Nikhil and Yu, Bin},
  booktitle={Forty-first International Conference on Machine Learning}
}

@inproceedings{tian2024hydralora,
  title={HydraLoRA: An Asymmetric LoRA Architecture for Efficient Fine-Tuning},
  author={Tian, Chunlin and Shi, Zhan and Guo, Zhijiang and Li, Li and Xu, Chengzhong},
  booktitle={Advances in Neural Information Processing Systems (NeurIPS)},
  year={2024}
}

@article{zhang2023lora,
  title={Lora-fa: Memory-efficient low-rank adaptation for large language models fine-tuning},
  author={Zhang, Longteng and Zhang, Lin and Shi, Shaohuai and Chu, Xiaowen and Li, Bo},
  journal={arXiv preprint arXiv:2308.03303},
  year={2023}
}

@inproceedings{mercea2024time,
  title={Time-Memory-and Parameter-Efficient Visual Adaptation},
  author={Mercea, Otniel-Bogdan and Gritsenko, Alexey and Schmid, Cordelia and Arnab, Anurag},
  booktitle={Proceedings of the IEEE/CVF Conference on Computer Vision and Pattern Recognition},
  pages={5536--5545},
  year={2024}
}

@book{brandwood2012fourier,
  title={Fourier transforms in radar and signal processing},
  author={Brandwood, David},
  year={2012},
publisher={Artech House}
}

@article{conrad2017sparse,
  title={Sparse Proteomics Analysis--a compressed sensing-based approach for feature selection and classification of high-dimensional proteomics mass spectrometry data},
  author={Conrad, Tim OF and Genzel, Martin and Cvetkovic, Nada and Wulkow, Niklas and Leichtle, Alexander and Vybiral, Jan and Kutyniok, Gitta and Sch{\"u}tte, Christof},
  journal={BMC Bioinformatics},
  volume={18},
  pages={1--20},
  year={2017}
}

@inproceedings{mevenkamp2016variational,
  title={Variational multi-phase segmentation using high-dimensional local features},
  author={Mevenkamp, Niklas and Berkels, Benjamin},
  booktitle={WACV},
  year={2016}
}

@article{zhai2019large,
  title={A large-scale study of representation learning with the visual task adaptation benchmark},
  author={Zhai, Xiaohua and Puigcerver, Joan and Kolesnikov, Alexander and Ruyssen, Pierre and Riquelme, Carlos and Lucic, Mario and Djolonga, Josip and Pinto, Andre Susano and Neumann, Maxim and Dosovitskiy, Alexey and others},
  journal={arXiv preprint arXiv:1910.04867},
  year={2019}
}

@inproceedings{van2015building,
  title={Building a bird recognition app and large scale dataset with citizen scientists: The fine print in fine-grained dataset collection},
  author={Van Horn, Grant and Branson, Steve and Farrell, Ryan and Haber, Scott and Barry, Jessie and Ipeirotis, Panos and Perona, Pietro and Belongie, Serge},
  booktitle={Proceedings of the IEEE conference on computer vision and pattern recognition},
  pages={595--604},
  year={2015}
}

@inproceedings{nilsback2008automated,
  title={Automated flower classification over a large number of classes},
  author={Nilsback, Maria-Elena and Zisserman, Andrew},
  booktitle={2008 Sixth Indian conference on computer vision, graphics \& image processing},
  pages={722--729},
  year={2008},
  organization={IEEE}
}

@inproceedings{khosla2011novel,
  title={Novel dataset for fine-grained image categorization: Stanford dogs},
  author={Khosla, Aditya and Jayadevaprakash, Nityananda and Yao, Bangpeng and Li, Fei-Fei},
  booktitle={Proc. CVPR workshop on fine-grained visual categorization (FGVC)},
  volume={2},
  number={1},
  year={2011}
}

@inproceedings{gebru2017fine,
  title={Fine-grained car detection for visual census estimation},
  author={Gebru, Timnit and Krause, Jonathan and Wang, Yilun and Chen, Duyun and Deng, Jia and Fei-Fei, Li},
  booktitle={Proceedings of the AAAI Conference on Artificial Intelligence},
  volume={31},
  number={1},
  year={2017}
}

@inproceedings{he2023sensitivity,
  title={Sensitivity-aware visual parameter-efficient fine-tuning},
  author={He, Haoyu and Cai, Jianfei and Zhang, Jing and Tao, Dacheng and Zhuang, Bohan},
  booktitle={Proceedings of the IEEE/CVF International Conference on Computer Vision},
  pages={11825--11835},
  year={2023}
}

@inproceedings{dreambooth,
  title={Dreambooth: Fine tuning text-to-image diffusion models for subject-driven generation},
  author={Ruiz, Nataniel and Li, Yuanzhen and Jampani, Varun and Pritch, Yael and Rubinstein, Michael and Aberman, Kfir},
  booktitle={Proceedings of the IEEE/CVF Conference on Computer Vision and Pattern Recognition},
  pages={22500--22510},
  year={2023}
}

@inproceedings{Caron_Touvron_Misra_Jegou_Mairal_Bojanowski_Joulin_2021,   title={Emerging Properties in Self-Supervised Vision Transformers},  url={http://dx.doi.org/10.1109/iccv48922.2021.00951},  DOI={10.1109/iccv48922.2021.00951},  booktitle={2021 IEEE/CVF International Conference on Computer Vision (ICCV)},  author={Caron, Mathilde and Touvron, Hugo and Misra, Ishan and Jegou, Herve and Mairal, Julien and Bojanowski, Piotr and Joulin, Armand},  year={2021},  month={Oct},  language={en-US}  }

@article{Radford_Kim_Hallacy_Ramesh_Goh_Agarwal_Sastry_Amanda_Mishkin_Clark_etal._2021,   title={Learning Transferable Visual Models From Natural Language Supervision},  journal={Cornell University - arXiv,Cornell University - arXiv},  author={Radford, Alec and Kim, JongWook and Hallacy, Chris and Ramesh, A. and Goh, Gabriel and Agarwal, Sandhini and Sastry, Girish and Amanda, Askell and Mishkin, Pamela and Clark, Jack and Krueger, Gretchen and Sutskever, Ilya},  year={2021},  month={Feb},  language={en-US}  }

@inproceedings{Zhang_Isola_Efros_Shechtman_Wang_2018,   title={The Unreasonable Effectiveness of Deep Features as a Perceptual Metric},  url={http://dx.doi.org/10.1109/cvpr.2018.00068},  DOI={10.1109/cvpr.2018.00068},  booktitle={2018 IEEE/CVF Conference on Computer Vision and Pattern Recognition},  author={Zhang, Richard and Isola, Phillip and Efros, Alexei A. and Shechtman, Eli and Wang, Oliver},  year={2018},  month={Jun},  language={en-US}  }

@article{wah2011caltech,
  title={The caltech-ucsd birds-200-2011 dataset},
  author={Wah, Catherine and Branson, Steve and Welinder, Peter and Perona, Pietro and Belongie, Serge},
  year={2011},
  publisher={California Institute of Technology}
}

@inproceedings{Wang_Singh_Michael_Hill_Levy_Bowman_2018,   title={GLUE: A Multi-Task Benchmark and Analysis Platform for Natural Language Understanding},  url={http://dx.doi.org/10.18653/v1/w18-5446},  DOI={10.18653/v1/w18-5446},  booktitle={Proceedings of the 2018 EMNLP Workshop BlackboxNLP: Analyzing and Interpreting Neural Networks for NLP},  author={Wang, Alex and Singh, Amanpreet and Michael, Julian and Hill, Felix and Levy, Omer and Bowman, Samuel},  year={2018},  month={Jan},  language={en-US}  }

@inproceedings{guo2024learning,
  title={Pela: Learning parameter-efficient models with low-rank approximation},
  author={Guo, Yangyang and Wang, Guangzhi and Kankanhalli, Mohan},
  booktitle={Proceedings of the IEEE/CVF Conference on Computer Vision and Pattern Recognition},
  pages={15699--15709},
  year={2024}
}

@inproceedings{kopiczkovera,
  title={VeRA: Vector-based Random Matrix Adaptation},
  author={Kopiczko, Dawid Jan and Blankevoort, Tijmen and Asano, Yuki M},
  booktitle={The Twelfth International Conference on Learning Representations}
}

@inproceedings{zhaogalore,
  title={GaLore: Memory-Efficient LLM Training by Gradient Low-Rank Projection},
  author={Zhao, Jiawei and Zhang, Zhenyu and Chen, Beidi and Wang, Zhangyang and Anandkumar, Anima and Tian, Yuandong},
  booktitle={Forty-first International Conference on Machine Learning}
}

@article{lian2022scaling,
  title={Scaling \& shifting your features: A new baseline for efficient model tuning},
  author={Lian, Dongze and Zhou, Daquan and Feng, Jiashi and Wang, Xinchao},
  journal={Advances in Neural Information Processing Systems},
  volume={35},
  pages={109--123},
  year={2022}
}

@inproceedings{chen2024conv,
  title={Conv-adapter: Exploring parameter efficient transfer learning for convnets},
  author={Chen, Hao and Tao, Ran and Zhang, Han and Wang, Yidong and Li, Xiang and Ye, Wei and Wang, Jindong and Hu, Guosheng and Savvides, Marios},
  booktitle={Proceedings of the IEEE/CVF Conference on Computer Vision and Pattern Recognition},
  pages={1551--1561},
  year={2024}
}

@inproceedings{peng2024parameter,
  title={Parameter efficient fine-tuning via cross block orchestration for segment anything model},
  author={Peng, Zelin and Xu, Zhengqin and Zeng, Zhilin and Xie, Lingxi and Tian, Qi and Shen, Wei},
  booktitle={Proceedings of the IEEE/CVF Conference on Computer Vision and Pattern Recognition},
  pages={3743--3752},
  year={2024}
}

@inproceedings{han20232,
  title={E 2 VPT: An Effective and Efficient Approach for Visual Prompt Tuning},
  author={Han, Cheng and Wang, Qifan and Cui, Yiming and Cao, Zhiwen and Wang, Wenguan and Qi, Siyuan and Liu, Dongfang},
  booktitle={2023 IEEE/CVF International Conference on Computer Vision (ICCV)},
  pages={17445--17456},
  year={2023},
  organization={IEEE}
}

@inproceedings{yin20231,
  title={1\% vs 100\%: Parameter-efficient low rank adapter for dense predictions},
  author={Yin, Dongshuo and Yang, Yiran and Wang, Zhechao and Yu, Hongfeng and Wei, Kaiwen and Sun, Xian},
  booktitle={Proceedings of the IEEE/CVF Conference on Computer Vision and Pattern Recognition},
  pages={20116--20126},
  year={2023}
}

@inproceedings{tu2023visual,
  author={Tu, Cheng-Hao and Mai, Zheda and Chao, Wei-Lun},
  booktitle={Proceedings of the IEEE/CVF Conference on Computer Vision and Pattern Recognition},
  pages={7725--7735},
  year={2023}
}

@inproceedings{he2022masked,
  title={Masked autoencoders are scalable vision learners},
  author={He, Kaiming and Chen, Xinlei and Xie, Saining and Li, Yanghao and Doll{\'a}r, Piotr and Girshick, Ross},
  booktitle={Proceedings of the IEEE/CVF conference on computer vision and pattern recognition},
  pages={16000--16009},
  year={2022}
}

@inproceedings{kirillov2023segment,
  title={Segment anything},
  author={Kirillov, Alexander and Mintun, Eric and Ravi, Nikhila and Mao, Hanzi and Rolland, Chloe and Gustafson, Laura and Xiao, Tete and Whitehead, Spencer and Berg, Alexander C and Lo, Wan-Yen and others},
  booktitle={Proceedings of the IEEE/CVF International Conference on Computer Vision},
  pages={4015--4026},
  year={2023}
}

@inproceedings{dosovitskiy2020image,
  title={An Image is Worth 16x16 Words: Transformers for Image Recognition at Scale},
  author={Dosovitskiy, Alexey and Beyer, Lucas and Kolesnikov, Alexander and Weissenborn, Dirk and Zhai, Xiaohua and Unterthiner, Thomas and Dehghani, Mostafa and Minderer, Matthias and Heigold, Georg and Gelly, Sylvain and others},
  booktitle={International Conference on Learning Representations},
  year={2020}
}

@article{zhang2025zookt,
  title={ZooKT: Task-adaptive knowledge transfer of Model Zoo for few-shot learning},
  author={Zhang, Baoquan and Shan, Bingqi and Li, Aoxue and Luo, Chuyao and Ye, Yunming and Li, Zhenguo},
  journal={Pattern Recognition},
  volume={158},
  pages={110960},
  year={2025},
  publisher={Elsevier}
}

@inproceedings{zhang2024codebook,
  title={Codebook Transfer with Part-of-Speech for Vector-Quantized Image Modeling},
  author={Zhang, Baoquan and Wang, Huaibin and Luo, Chuyao and Li, Xutao and Liang, Guotao and Ye, Yunming and Qi, Xiaochen and He, Yao},
  booktitle={Proceedings of the IEEE/CVF Conference on Computer Vision and Pattern Recognition},
  pages={7757--7766},
  year={2024}
}

@inproceedings{lin2023vision,
  title={Vision transformers are parameter-efficient audio-visual learners},
  author={Lin, Yan-Bo and Sung, Yi-Lin and Lei, Jie and others},
  booktitle={Proceedings of the IEEE/CVF Conference on Computer Vision and Pattern Recognition},
  pages={2299--2309},
  year={2023}
}

@inproceedings{zhang2024gradient,
  title={Gradient-based Parameter Selection for Efficient Fine-Tuning},
  author={Zhang, Zhi and Zhang, Qizhe and Gao, Zijun and Zhang, Renrui and Shutova, Ekaterina and Zhou, Shiji and Zhang, Shanghang},
  booktitle={Proceedings of the IEEE/CVF Conference on Computer Vision and Pattern Recognition},
  pages={28566--28577},
  year={2024}
}

@inproceedings{liudora,
  title={DoRA: Weight-Decomposed Low-Rank Adaptation},
  author={Liu, Shih-yang and Wang, Chien-Yi and Yin, Hongxu and Molchanov, Pavlo and Wang, Yu-Chiang Frank and Cheng, Kwang-Ting and Chen, Min-Hung},
  booktitle={Forty-first International Conference on Machine Learning}
}

@inproceedings{zhu2024asymmetry,
  title={Asymmetry in Low-Rank Adapters of Foundation Models},
  author={Zhu, Jiacheng and Greenewald, Kristjan and Nadjahi, Kimia and Borde, Haitz Saez De Ocariz and Gabrielsson, Rickard and Choshen, Leshem and Ghassemi, Marzyeh and Yurochkin, Mikhail and Solomon, Justin},
  booktitle={International Conference on Learning Representations},
  year={2024}
}

@article{HURKENS20041,
title = {On the nearest neighbor rule for the traveling salesman problem},
journal = {Operations Research Letters},
volume = {32},
number = {1},
pages = {1-4},
year = {2004},
issn = {0167-6377},
doi = {https://doi.org/10.1016/S0167-6377(03)00093-2},
url = {https://www.sciencedirect.com/science/article/pii/S0167637703000932},
author = {Cor A.J. Hurkens and Gerhard J. Woeginger}
}

@inproceedings{
      chen2024quanta,
      title={Quan{TA}: Efficient High-Rank Fine-Tuning of {LLM}s with Quantum-Informed Tensor Adaptation},
      author={Zhuo Chen and Rumen Dangovski and Charlotte Loh and Owen M Dugan and Di Luo and Marin Soljacic},
      booktitle={The Thirty-eighth Annual Conference on Neural Information Processing Systems},
      year={2024},
}

@inproceedings{NEURIPS2024_75008a0f,
 author = {Wu, Zhengxuan and Arora, Aryaman and Wang, Zheng and Geiger, Atticus and Jurafsky, Dan and Manning, Christopher D. and Potts, Christopher},
 booktitle = {Advances in Neural Information Processing Systems},
 pages = {63908--63962},
 title = {ReFT: Representation Finetuning for Language Models},
 volume = {37},
 year = {2024}
}

@inproceedings{
huang2025hira,
title={Hi{RA}: Parameter-Efficient Hadamard High-Rank Adaptation for Large Language Models},
author={Qiushi Huang and Tom Ko and Zhan Zhuang and Lilian Tang and Yu Zhang},
booktitle={The Thirteenth International Conference on Learning Representations},
year={2025},
}

@inproceedings{
he2025smt,
title={{SMT}: Fine-Tuning Large Language Models with Sparse Matrices},
author={Haoze He and Juncheng B Li and Xuan Jiang and Heather Miller},
booktitle={The Thirteenth International Conference on Learning Representations},
year={2025},
}

@inproceedings{
woo2025paca,
title={Pa{CA}: Partial Connection Adaptation for Efficient Fine-Tuning},
author={Sunghyeon Woo and Sol Namkung and Sunwoo Lee and Inho Jeong and Beomseok Kim and Dongsuk Jeon},
booktitle={The Thirteenth International Conference on Learning Representations},
year={2025},
}

@inproceedings{
liao2025hmora,
title={{HM}o{RA}: Making {LLM}s More Effective with Hierarchical Mixture of Lo{RA} Experts},
author={Mengqi Liao and Wei Chen and Junfeng Shen and Shengnan Guo and Huaiyu Wan},
booktitle={The Thirteenth International Conference on Learning Representations},
year={2025},
}

@inproceedings{wang2023self,
  title={Self-instruct: Aligning language models with self-generated instructions},
  author={Wang, Yizhong and Kordi, Yeganeh and Mishra, Swaroop and Liu, Alisa and Smith, Noah A and Khashabi, Daniel and Hajishirzi, Hannaneh},
  booktitle={Proceedings of the 61st annual meeting of the association for computational linguistics (volume 1: long papers)},
  pages={13484--13508},
  year={2023}
}

@article{zheng2023judging,
  title={Judging llm-as-a-judge with mt-bench and chatbot arena},
  author={Zheng, Lianmin and Chiang, Wei-Lin and Sheng, Ying and Zhuang, Siyuan and Wu, Zhanghao and Zhuang, Yonghao and Lin, Zi and Li, Zhuohan and Li, Dacheng and Xing, Eric and others},
  journal={Advances in neural information processing systems},
  volume={36},
  pages={46595--46623},
  year={2023}
}

@article{chiang2023vicuna,
  title={Vicuna: An open-source chatbot impressing gpt-4 with 90\%* chatgpt quality},
  author={Chiang, Wei-Lin and Li, Zhuohan and Lin, Ziqing and Sheng, Ying and Wu, Zhanghao and Zhang, Hao and Zheng, Lianmin and Zhuang, Siyuan and Zhuang, Yonghao and Gonzalez, Joseph E and others},
  journal={See https://vicuna. lmsys. org (accessed 14 April 2023)},
  volume={2},
  number={3},
  pages={6},
  year={2023}
}
